\documentclass[onecolumn]{openhelix}
\usepackage{graphicx}
\usepackage{amsmath}
\usepackage{amssymb}
\usepackage{amsfonts}
\usepackage{tabularx}
\usepackage{float}
\usepackage{algorithm}
\usepackage{algorithmic}
\usepackage{utfsym}
\usepackage{array}
\usepackage{booktabs}
\usepackage{makecell}
\usepackage{wrapfig}
\usepackage{xspace}

\usepackage[table]{xcolor}
\definecolor{newyellow}{HTML}{FFD94D}
\definecolor{newgrey}{HTML}{7F7F7F}
\definecolor{newpink}{HTML}{FBCDF4}
\definecolor{realworldoft}{HTML}{029533}
\definecolor{realworldsf}{HTML}{8CC46A}
\definecolor{realworldour}{HTML}{FFC715}
\newcommand{\method}{\texttt{Fast-dVLA}}
\usepackage[most]{tcolorbox}
\tcbuselibrary{theorems}
\newtcbtheorem{finding}{Finding}{
  colback=openhelixred!5,
  colframe=openhelixred!60!black,
  fonttitle=\bfseries,
}{find}

\title{Fast-dVLA: Accelerating Discrete Diffusion VLA to Real-Time Performance}

\author[1*]{Wenxuan Song}
\author[1*]{Jiayi Chen}
\author[2,3*]{Shuai Chen}
\author[1]{Jingbo Wang}
\author[5,6]{Pengxiang Ding}
\author[5,6]{Han Zhao}
\author[1]{Yikai Qin}
\author[1]{Xinhu Zheng}
\author[2]{Donglin Wang}
\author[4\dagger]{Yan Wang}
\author[1\dagger]{Haoang Li}

\affiliation[1]{The Hong Kong University of Science and Technology (Guangzhou)}
\affiliation[2]{ShanghaiTech University}
\affiliation[3]{Shanghai Institute of Technical Physics, CAS}
\affiliation[4]{AIR, Tsinghua University}
\affiliation[5]{Westlake University}
\affiliation[6]{Zhejiang University}

\contribution[*]{Equal Contribution}
\contribution[\dagger]{Corresponding Author}

\abstract{
This paper proposes a novel approach to address the challenge that pretrained VLA models often fail to effectively improve performance and reduce adaptation costs during standard supervised finetuning (SFT). Some advanced finetuning methods with auxiliary training objectives can improve performance and reduce the number of convergence steps. However, they typically incur significant computational overhead due to the additional losses from auxiliary tasks. To simultaneously achieve the enhanced capabilities of auxiliary training with the simplicity of standard SFT, we decouple the two objectives of auxiliary task training within the parameter space, namely, enhancing general capabilities and fitting task-specific action distributions. To deliver this goal, we only need to train the model to converge on a small-scale task set using two distinct training strategies. The difference between the resulting model parameters can then be interpreted as \textit{capability vectors} provided by auxiliary tasks. These vectors are then merged with pretrained parameters to form a capability-enhanced meta model. Moreover, when standard SFT is augmented with a lightweight orthogonal regularization loss, the merged model attains performance comparable to auxiliary finetuned baselines with reduced computational overhead. Experimental results demonstrate that this approach is highly effective across diverse robot tasks. Project page: \url{https://chris1220313648.github.io/Fast-dVLA/}
}

\correspondence{%
\begin{tabular}[t]{@{}l@{\ }l@{}}
Wenxuan Song at & \email{songwenxuan0115@gmail.com}\\
\end{tabular}%
}

\begin{document}

\maketitle

\section{Introduction}

%VLA往往从大规模机器人数据集上进行训练，将多模态感知转化为可执行的robotic control，表现出一定的语言跟随和视觉泛化能力，在当前的robotic foundation model研究中占据主导地位。代表性的VLA~\citep{pi0, gr00t}采用vlm + action head的架构，其中vlm进行多模态理解，action head为diffusion 或 flow-matching架构，接收vlm处理的information并输出连续的控制信号。最近，基于dllm架构的discrete diffusion vlas~\citep{discrete diffusion vla, dvla, llada-vla, ud-vla, dreamvla} have emerged as a promising challenger to existing VLA architectures. 得益于离散的动作输出空间，他们将感知和控制的任务统一在一个模型中，从而实现了无缝的信息传输，并得到了comparative的性能。
Vision–Language–Action (VLA)~\citep{yan2026svam,cui2025openhelix,intelligence2025pi_,song2025accelerating,kim2024openvla} models are typically trained on large-scale robotic datasets to map multimodal perception into executable robotic control, exhibiting language following and visual generalization capabilities, and have become a dominant paradigm in current research on robotic foundation models. Representative VLAs~\citep{Pi0, gr00t, liu2025hybridvla,zhong2026dualcot} adopt a flow-matching architecture, where the VLM performs multimodal understanding and the FM action head takes the processed representations as input and outputs continuous control signals.
Recently, discrete diffusion VLAs (dVLAs) based on the diffusion Large Language Models (dLLMs)~\citep{liang2025discrete, wen2025dvla, wen2025llada, chen2025unified, ye2025dream} have emerged as a promising challenger to existing VLA architectures. 
These models output actions in a parallel, iterative denoising manner, while not relying on the flow-matching head.
%因此相对于FM架构展现出更强的统一模态对齐和理解能力，同时更好的保留了VLM的预训练知识~\citep{ddvla}，
Therefore, compared with flow-matching architectures, they demonstrate inherent advantages in unified multimodal alignment and understanding~\citep{chen2025unified, wen2025dvla}, while better preserving the pretrained knowledge of VLMs~\citep{liang2025discrete}.

% By operating in a unified discrete action space, these models unify perception and control within a single model, enabling seamless information flow between modalities and actions, and achieving competitive performance.
\begin{figure*}[t]
    \centering
    \includegraphics[width=0.98\linewidth]{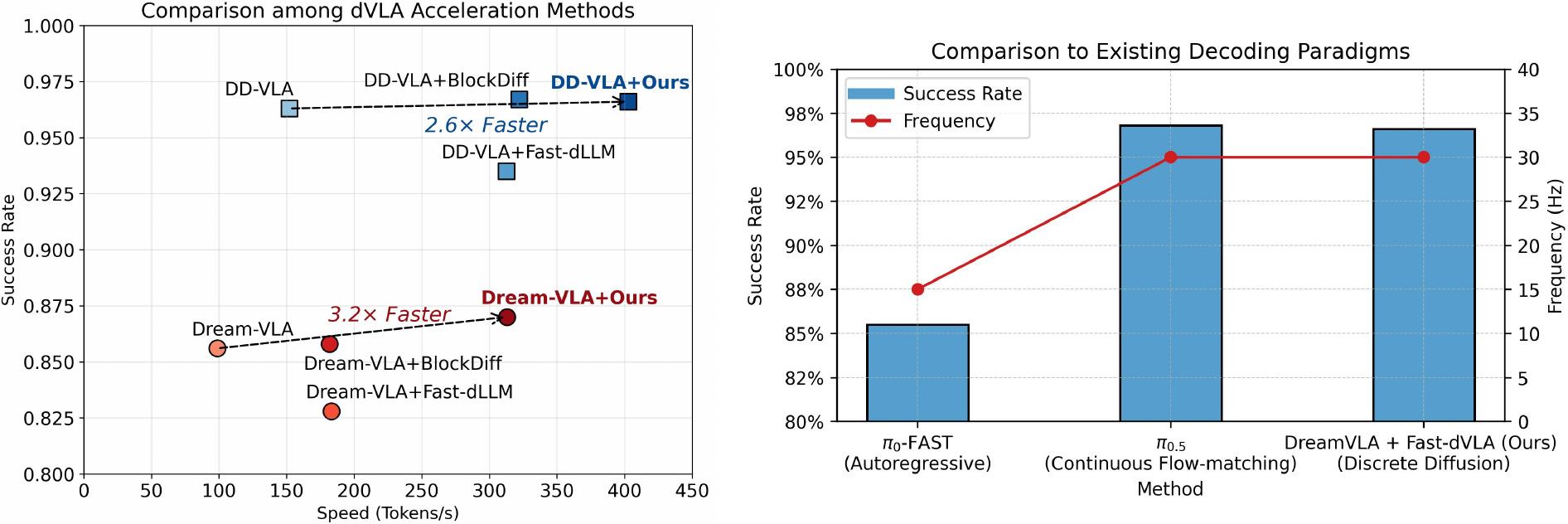}
    \caption{\textbf{Speed/Success Rate trade-off.}
    \textbf{Left (Intra-comparison):} Compared to other acceleration strategies for discrete diffusion VLAs (dVLAs), DD-VLA~\citep{liang2025discrete} and Dream-VLA~\citep{yedreamVLA}, our \method~achieves a favorable success rate and speed.
    Here, BlockDiff denotes block diffusion~\citep{arriola2025block}.
    \textbf{Right (Inter-comparison):} Our \method~surpasses autoregressive methods, \textit{i.e.}, $\pi_0$-FAST~\citep{pertsch2025fast}.
    It also reaches parallel performance and inference frequency with state-of-the-art (SOTA) continuous flow-matching methods, \textit{i.e.}, $\pi_{0.5}$~\citep{intelligence2025pi_}, while maintaining several inherent advantages of dVLAs.
    We report metrics on LIBERO~\citep{liu2023libero}.}
    \label{fig:abs_teaser}
\end{figure*}

\begin{figure*}[t]  
    \centering
    \includegraphics[width=0.98\linewidth]{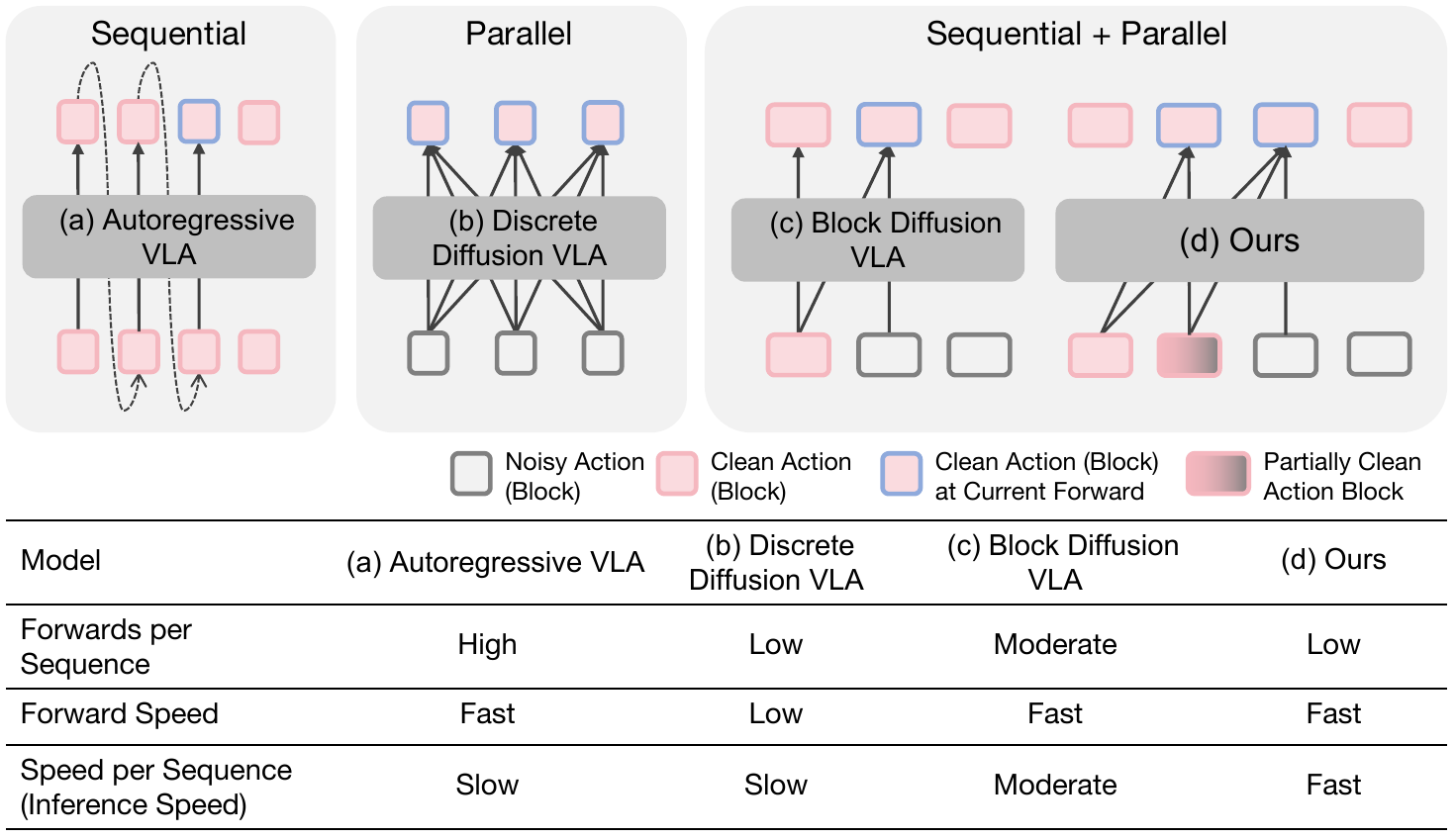} 
    \caption{\textbf{Comparison among discrete decoding paradigms.} Here, \textit{Forward per Sequence} denotes the needed forward numbers for a full sequence output, \textit{Forward Speed} denotes the decoding speed for each forward, and \textit{Speed per Sequence (i.e., Inference Speed)} denotes the decoding speed for the full sequence output.
    \textbf{Our \method~requires significantly fewer forward passes and executes each pass efficiently, resulting in substantially faster inference.}
    }
    \label{fig:teaser} 
\end{figure*}

%然而，目前的dVLA面临根本性问题：他们的推理速度非常慢，输出频率约为1hz，这与真实世界机器人任务的实时要求（10hz）相差甚远，这导致了他们脱离了真实世界的可用性。如图1所示，虽然相对于discrete autoregressive (AR) vla, dVLA通过parallel计算显著的减少了推理一个完整序列所需的forward次数，然而，双向注意力的特性使其无法利用前面token的kv cache，从而导致了非常低的forward speed。一个可能的改进方式是将纯粹的diffusion process重塑为AR + Diffusion，直观的方案是employ block diffusion~\citep{block diffusion}, 他一次并行解码一个block中的action token，并在完成整个block的解码后存储kv cache，并切换到下一个block继续解码。这种方案同时保证了一定的kv cache利用率和并行解码带来的更少的forward过程，因此得到了moderate的推理速度。Yet, it precludes the interblock parallelism, a crucial factor for efficient inference.
However, current dVLAs still suffer from a fundamental limitation. Their inference speed is slow, with an execution frequency that is far below the real-time requirements of physical robotic systems (typically around 30~Hz). This large gap substantially limits their practical applicability in real-world settings. As illustrated in \Cref{fig:teaser}, although dVLA significantly reduces the number of forward passes required to generate a complete action sequence compared to discrete autoregressive (AR) VLA (\Cref{fig:teaser}~(a)) by enabling parallel decoding, its bidirectional attention mechanism prevents the reuse of key--value (KV) caches from previously generated tokens, resulting in a very low per-pass forward efficiency (\Cref{fig:teaser}~(b)).

% A natural direction to mitigate this issue is to restructure the purely parallel decoding process into a partially sequential decoding process, \textit{i.e.}, a hybrid AR--diffusion framework. 
% We empirically observe that, although with bidirectional attention, dVLAs still follow a left-to-right decoding pattern. 
% This block-wise AR decoding behavior suggests that a finetuned bidirectional discrete diffusion model can be explicitly forced to follow block diffusion~\citep{arriola2025block} (\Cref{fig:teaser}~(c)).
% % This block-wise AR decoding behavior naturally motivates block diffusion.  with block attention
% It decodes tokens within a block in parallel, caches the corresponding KV states once the block is completed, and then proceeds to the next block in an AR fashion. This design achieves moderate inference speed by balancing partial KV-cache reuse with intra-block parallel decoding. However, it inherently disallows inter-block parallelism, which is critical for achieving high throughput and low-latency inference.

%为了探索利用KV cache的可能性，我们观察了dVLA的动作解码顺序
To explore the feasibility of leveraging KV cache, we investigate the action decoding order in dVLAs (\Cref{fig:trace}).
We observe that, although with bidirectional attention, dVLAs still follow a left-to-right decoding pattern. 
This block-wise decoding behavior suggests that a promising direction is to apply block diffusion~\citep{arriola2025block} (\Cref{fig:teaser}~(c)), which natively trains dVLAs with block-wise attention,
decodes a block of action tokens in parallel, caches the corresponding KV states after completing the block, and then proceeds to the next block in an AR manner. This design achieves a moderate inference speed by balancing partial KV cache reuse with intra-block parallel decoding. However, it inherently precludes inter-block parallelism, which is a crucial factor for achieving high-throughput and low-latency inference.

This paper proposes \method, a novel block-wise diffusion 
% VLA 
strategy
that achieves the first breakthrough in accelerating dVLAs to a real-time regime. 
% Conceptually, we aim to embrace block-wise sequential generation to facilitate KV cache utilization, yet reject the dilemma that the decoding of subsequent blocks must wait for preceding blocks to be fully denoised. 
Conceptually, we exploit block-wise sequential generation for KV cache utilization, while removing the requirement that later blocks wait for earlier ones to finish denoising.
Concretely, we treat the full action token sequence at each timestep (\textit{i.e.}, the dimensionality of the actions) and its multiples as an action block.
Then, \method~learns to denoise a sequence of blocks with monotonically increasing mask ratios in parallel. 
Naturally, preceding blocks can finish before subsequent ones, allowing their KV states to be cached for subsequent computations. 
Note that we constrain the attention to be block-wise causal to ensure the KV cache remains unchanged. 
For training efficiency, inspired by \citep{wang2026diffusion}, we distill \method~from finetuned dVLAs with bidirectional attention using an asymmetric distillation loss. 
During inference, we design a pipelined parallel decoding algorithm that enables inter-block parallelism with varying noise levels across blocks.
% cjy 我们在dvla的代表工作dreamvla, discrete difussion vla以及同时生成视觉cot辅助动作生成的UD-vla进行了实验验证，我们的方法展现一致性的加速提升，在各个CALVIN LIBERO SIMPLER以及real-world是实验上达到2.8~4.1的加速比，大幅度提高推理效率的同时保证动作性能几乎不变,甚至由于保持动作块之间的因果性，减少了模型的困惑性，在部分任务上性能得到了提升.

We conduct extensive experiments on representative dVLA models, including Dream-VLA~\citep{yedreamVLA}, Discrete Diffusion VLA (DD-VLA)~\citep{liang2025discrete}, and UD-VLA~\citep{chen2025unified} across CALVIN~\citep{mees2022calvin}, LIBERO~\citep{liu2023libero}, and SIMPLER~\citep{li24simpler} benchmarks. 
\Cref{fig:abs_teaser} shows that our method consistently achieves 2.8$\times$–4.1$\times$ speedup while preserving action performance, which is superior to other dVLA paradigms.
Lastly, diverse real-world tasks demonstrate the dynamic capability and working efficiency in the application.
% and, in some cases, even improves performance by maintaining causal consistency across action blocks and effectively reducing model perplexity~\citep{ni2026flexibility}.

Our contributions are summarized as follows:

\begin{itemize}

% \item We identify a fundamental bottleneck in existing discrete diffusion VLA (dVLA) models: although parallel decoding reduces the number of forward passes, bidirectional attention prevents effective KV cache reuse, leading to extremely low real-time inference frequency.
\item We reveal an implicit block-wise AR decoding tendency in the fully bidirectional dVLA, motivating an AR--diffusion hybrid denoising process.
\item We propose \method, which leverages block-wise diffusion with a corresponding attention pattern to allow KV cache reuse, while allowing inter-block parallelism through diffusion forcing. 
\item Following the observation, we apply an asymmetric distillation for efficient training, and a pipelined parallel decoding for real-time inference.
\item Extensive experiments on CALVIN, LIBERO, and SIMPLER demonstrate up to 4.1$\times$ acceleration over existing dVLA models, while maintaining SOTA-level success rates.
Moreover, the results in diverse real-world tasks demonstrate the dynamic capability and working efficiency in the application.

\end{itemize}

%考虑到dVLA自然的作为unified vla model支持同时进行future image generation and action prediction，我们将我们的方案从单一模态输出扩展至多模态输出。其中teacher model的image generation和action prediction采用双向注意力。考虑到image token的数量远大于action token的数量，两种类型的token使用不同大小的block size。
% Motivated by the fact that dVLA naturally serves as a unified VLA capable of jointly generating future images and predicting actions, we extend our approach from a single-modality output to a multimodal output setting. In the teacher model, image generation and action prediction are implemented using bidirectional attention to ensure an interactive information transmission.
% Since the number of image tokens is substantially larger than that of action tokens, we adopt different block sizes for the two token types.
\section{Preliminary: Discrete Diffusion VLA (dVLA)}

Discrete Diffusion VLA (\emph{e.g.}, \textbf{Dream-VLA}\citep{yedreamVLA} and \textbf{DD-VLA}~\citep{liang2025discrete}) output discrete action tokens, obtained either by uniform bins~\citep{kim2024openvla} or by quantized tokenizers~\citep{pertsch2025fast}, instead of operating directly on continuous controls.
% Discrete diffusion VLAs 通过离散动作为256个bin分位或者使用fast 动作编码器根据频域信息将动作编码离散动作 rather than continuous controls.
Actions are represented as a length-$L$ discrete token sequence
$\mathbf{a}_0=(a_{0}^i,\dots, a_{0}^L)$, where each token $a_{0}^i$ corresponds to discrete low-level robot actions and a special mask token $\mathrm{M}$ is added to the vocabulary to enable diffusion-style corruption.

The forward diffusion process randomly replaces a subset of action tokens with $\mathrm{M}$ according to a time-dependent mask ratio, independently across positions.
The reverse process learns to recover masked tokens conditioned on the unmasked context and multimodal inputs $\mathbf{c}$ (\emph{e.g.}, language and visual observations).
At each denoising step, unmasked tokens are copied unchanged, while masked positions are predicted from a categorical distribution parameterized by the model.

During training, a mask ratio $\gamma_t\in(0,1]$ is sampled, and the corresponding action tokens are replaced by $\mathrm{M}$ to obtain a corrupted sequence $\tilde{\mathbf{a}}_t$.
The model is then trained to reconstruct the original tokens using cross-entropy loss computed only on masked positions:
\begin{equation}
\mathcal{L}_{\text{act}}(\theta)
= - \sum_{i \in \mathcal{M}_{\gamma_t}}
\log p_\theta\!\left(a_{0}^i \mid \tilde{\mathbf{a}}_t, \mathbf{c}\right),
\label{eq:masked_ce}
\end{equation}
where $\mathcal{M}_{\gamma_t}$ denotes the set of masked positions.
This objective preserves the core corruption–denoising principle of discrete diffusion while enabling efficient training with standard discrete VLA architectures.

% Unified dVLA在dVLA的基础上，将未来帧预测引入到训练和推理过程中，具体的，我们通过vq-vae编码编码未来图像成离散的token序列$\mathbf{v}_0=(v_{0,1},\dots,v_{0,L_v})$,将其与动作序列拼接成y0=(v_0,a_0),在forward过程推理时，我们共同扩散生成未来帧以及动作，实现理解和生成的相互裨益，训练损失同样采用ross-entropy loss：
\textbf{UD-VLA}~\citep{chen2025unified} extends this framework to unified VLA models\citep{univla} by incorporating future visual prediction.
Specifically, future image observations are encoded into a discrete token sequence
$\mathbf{v}_0=(v_{0}^1,\dots,v_{0}^{L})$ using a VQ-VAE~\citep{zheng2022movq} encoder, and concatenated with the action tokens to form a unified sequence. 
% $\mathbf{y}_0=(\mathbf{v}_0,\mathbf{a}_0)$.
The diffusion process is then applied jointly over visual and action tokens, allowing future visual reasoning and action generation to be learned in a unified manner.
% The model is optimized to reconstruct the original tokens with cross-entropy computed only on masked positions:
% \begin{equation}
% \mathcal{L}_{\text{uni}}(\theta)
% =
% -\sum_{r\in \mathcal{M}_{\gamma_t}}
% w \log p_\theta\!\left(y_{0}^r\mid \tilde{\mathbf{y}}_t,\mathbf{c}\right),
% \label{eq:unified_masked_ce_single}
% \end{equation}
% where $w$ assigns different loss weights to visual and action tokens, with visual tokens down-weighted to prevent dominance.

% The training objective similarly adopts a cross-entropy loss over masked tokens in both modalities, enabling mutual reinforcement between perception and action generation.

% 变量定义
% 前向加噪
% 反向去噪

\section{Method}
In this section, we first introduce an intriguing observation~(see~\Cref{sec:motivations}).
Then, we propose \method~to accelerate dVLA in a block-wise decoding manner.
Our method is built to support two key features: (1) a block-wise attention mechanism that enables the reuse of KV cache across denoising iterations~(see~\Cref{sec:kv}) and (2) a diffusion forcing denoising process that supports simultaneous decoding of blocks with different noising levels.
To efficiently train such models, we design an asymmetric distillation that starts from a pretrained bidirectional dVLA~(see~\Cref{sec:train}).
During inference, we design an inter-block parallel decoding schedule that balances inference speed and decoding reliability~(see~\Cref{sec:infer}).

% Together, these designs yield a stable, cache-efficient, and confidence-controlled decoding framework that significantly reduces computational overhead without sacrificing action generation quality.

\begin{figure}[t]  
    \centering
    \includegraphics[width=0.99\linewidth]{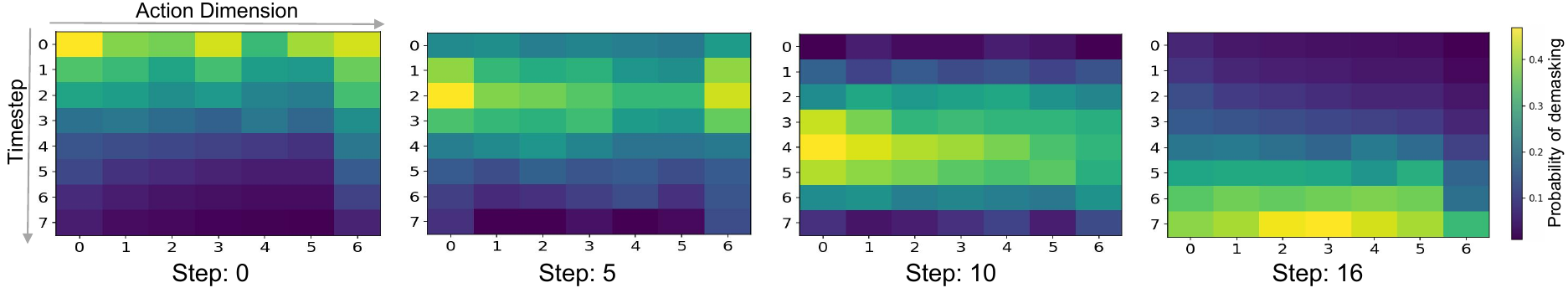} 
    \caption{Visualization of the decoding tendency of action tokens at different positions in Dream-VLA~\citep{yedreamVLA}.
    Brighter regions indicate higher decoding probability.
    Despite using bidirectional attention, the model exhibits a clear left-to-right decoding tendency such that action tokens at earlier temporal positions are typically decoded in earlier diffusion iterations.
    Overall, the decoding process reveals an implicit block-wise AR pattern.}

    \label{fig:trace} 
\end{figure}

\subsection{Motivation}
\label{sec:motivations}
As shown in~\Cref{fig:trace}, we record and visualize the decoding frequency at different positions during the denoising process of a representative dVLA (\textit{i.e.,} Dream-VLA). 
Interestingly, even though the dVLA employs bidirectional attention, the model still exhibits a strong left-to-right decoding pattern at a global level. In particular, action blocks that occur earlier in the temporal dimension tend to be decoded in earlier denoising iterations.
% This can be attributed to 1) 现有dVLA的backbone 是由ar llm初始化并训练成dllm的，所以还继承了自回归的性质 2)不同时间步的动作之间存在时序上的依赖。
This can be attributed to: 1) The backbone~\citep{ye2025dream} of existing dVLAs is typically initialized from an AR VLM and trained in a discrete diffusion manner, thereby retaining certain autoregressive characteristics. 2) Actions at different timesteps exhibit inherent temporal dependencies.
This block-wise AR decoding behavior suggests that \textit{a finetuned bidirectional dVLA can be directly forced to follow a block-diffusion decoding manner}.

\subsection{Critical Designs of Target Models}
\label{sec:kv}
\begin{figure*}[t]
    \centering
    \subfloat[Bidirectional Attention]{
        \includegraphics[width=0.46\linewidth]{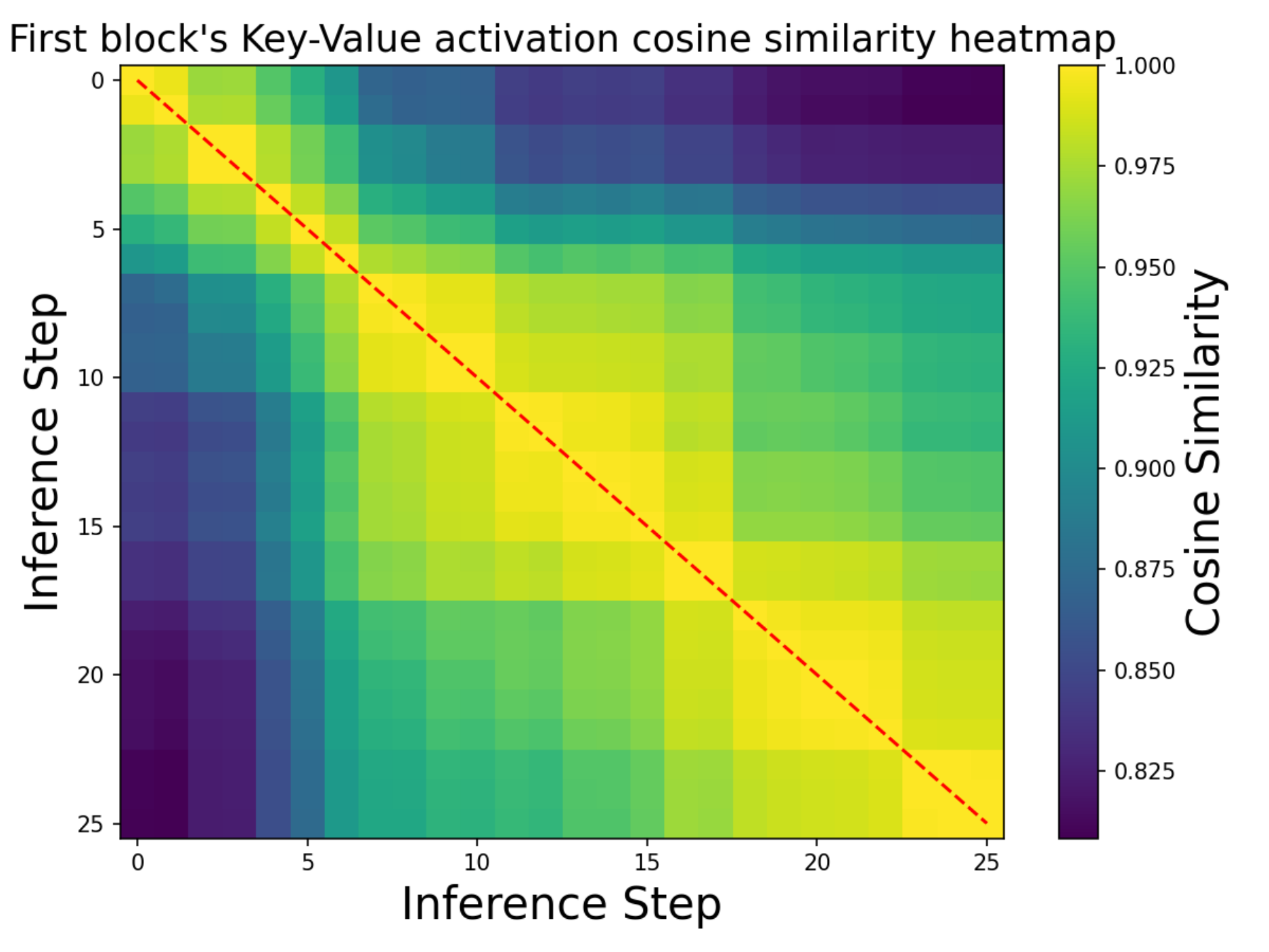}
        \label{fig:attn_full}
    }
    % \hfill
    \subfloat[Block-wise Attention]{
        \includegraphics[width=0.46\linewidth]{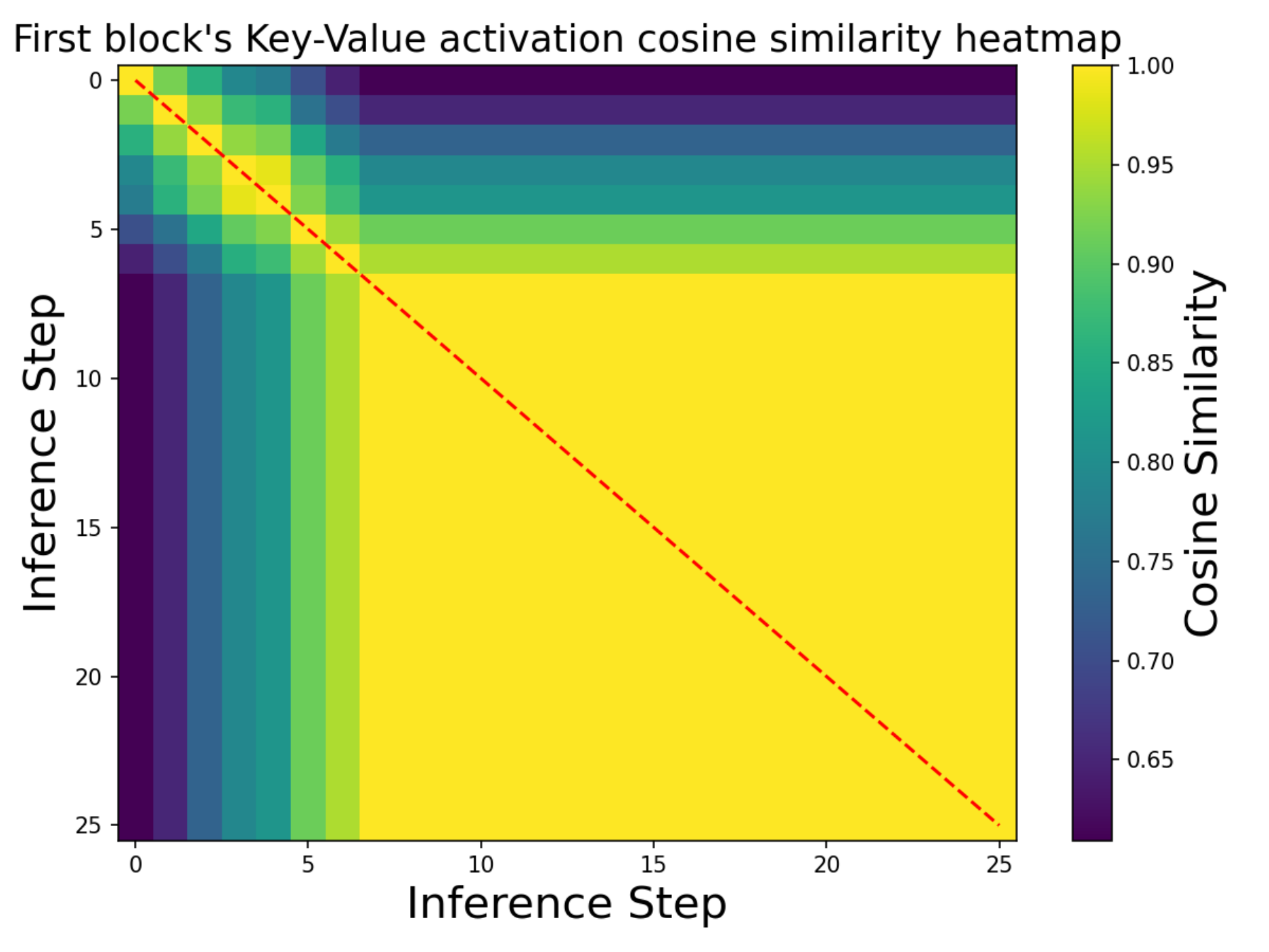}
        \label{fig:attn_block}
    }
    \caption{
\textbf{KV cache similarity across diffusion iterations under block-diffusion decoding.}
    We visualize the similarity of attention key--value (KV) states for the first action block across different denoising steps.
    \textbf{(a):}
    In native dVLA with bidirectional attention, the KV representations evolve across iterations, preventing effective reuse of cached states.
    \textbf{(b):}
    In contrast, after adapting dVLA to a block-wise attention architecture via asymmetric distillation, once all tokens in the first block are unmasked, the corresponding KV states remain fixed,
    enabling efficient KV cache reuse and substantially reducing the computational overhead in subsequent iterations.
    }
    \label{fig:compare_attn}
\end{figure*}

% 目前的Diffusion VLA生成部分或者全部的序列都采用双向注意力，这会导致每次迭代的过程中的kv cache会产生变化，无法复用kvcache进行加速，Block-Wise Decoding采用Block级的注意力，块内的kvcahe 在迭代过程中只受prefix token以及块内token的影响，一旦块内解码完成，后续解码过程kv值就会不变，所以可以复用给后续解码

\begin{wrapfigure}{r}{0.3\textwidth}
    \vspace{-0.2cm}
    \centering
    \includegraphics[width=0.98\linewidth]{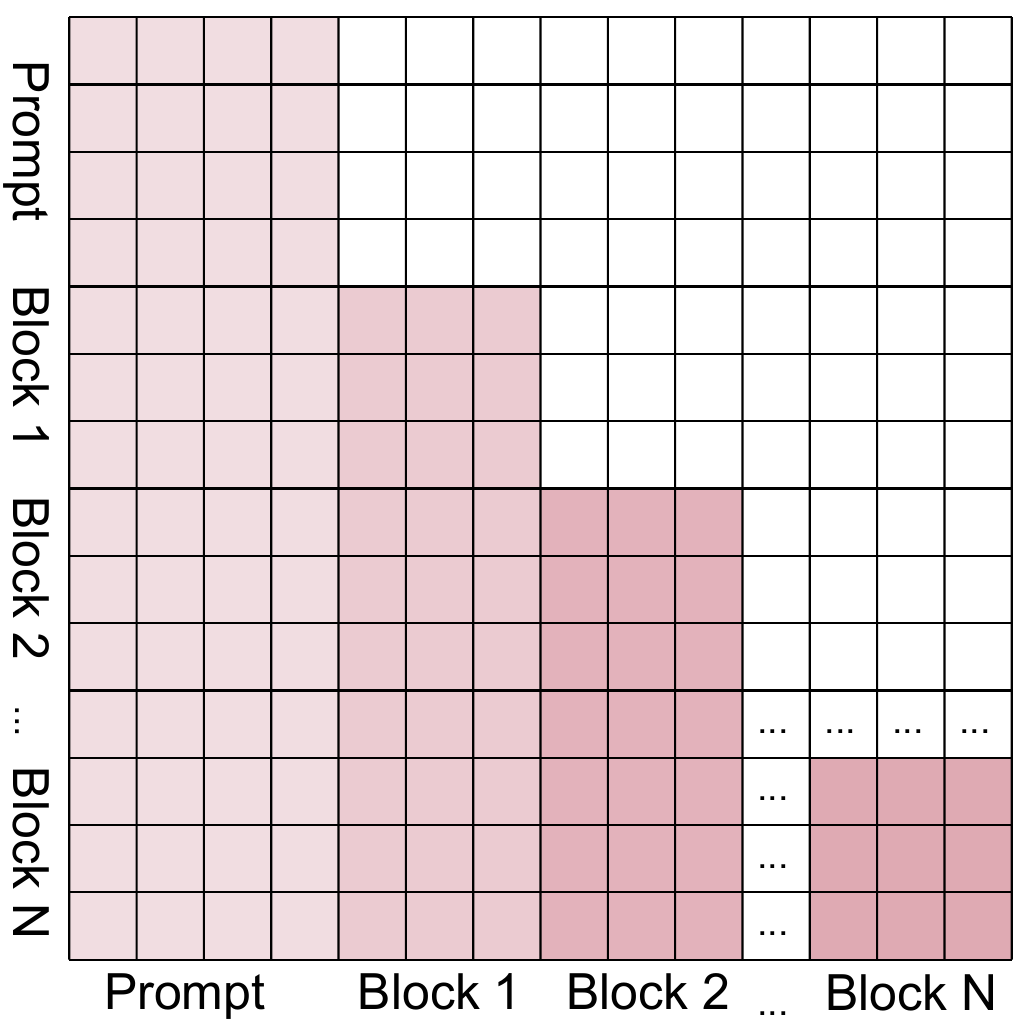} 
    \caption{\textbf{Block-wise Attention.} }
    \label{fig:attention} 
    \vspace{-0.2cm}
\end{wrapfigure}

\noindent
\textbf{Block-Wise Attention for Inter-block KV Cache Reusing.}
As shown in \Cref{fig:attn_full}, current dVLA~\citep{chen2025unified,liang2025discrete,yedreamVLA} models generate either partial or full sequences using bidirectional attention, which causes the Key-Value (KV) representations to vary at every denoising iteration. As a result, the conventional KV cache mechanism used in AR models cannot be directly reused to accelerate inference.
To address this limitation, we adopt a block diffusion decoding strategy (see~\Cref{fig:teaser}d) with block-wise attention (see~\Cref{fig:attention}), which bridges autoregressive decoding and discrete diffusion by interpolating between sequential dependency and parallel generation. 

Within each block, the KV representations are influenced only by the prefix tokens and the tokens inside the current block. 
As shown in \Cref{fig:attn_block}, once the decoding of a block is completed, the KV values of that block remain unchanged in subsequent steps, enabling effective cache reuse for the following decoding process.

% This design allows us to use a stable KV Cache under block-wise attention, significantly reducing redundant computation without modifying the underlying model architecture.
% 贴一个解码过程中的注意力热图，表示不同迭代步之间注意力的相关性，双向的会随迭代步数变化，而block difussion一旦块内的解码出来了就不会变了
% \begin{figure*}[t]  
%     \centering
%     \includegraphics[width=0.8\linewidth]{figure/attention.png} 
%     \caption{\textbf{example.} }
%     \label{fig:attention} 
% \end{figure*}

% \begin{figure*}[t]
%     \centering
%     \subfigure[Full bi-directional attention]{
%         \includegraphics[width=0.45\linewidth]{figure/dvla.pdf}
%     }
%     \hfill
%     \subfigure[Block-wise attention]{
%         \includegraphics[width=0.45\linewidth]{figure/blockdvla.pdf}
%     }
%     \caption{\textbf{Example.} Comparison of attention similarity patterns under different decoding strategies.}
%     \label{fig:attention}
% \end{figure*}

\noindent
\textbf{Diffusion Forcing for Inter-block Parallel Decoding.}
\label{sec:forcing}
Motivated by observations in mimic-video~\citep{pai2025mimic} that action tokens need not attend to the clean tokens from previous timesteps, we construct a progressively decaying noise sequence similar to diffusion forcing~\citep{chen2024diffusion,yin2025slow,li2026causal}.
% Let ${B_i} := \{(i-1)*k, \dots, i*k - 1\}$ denote the token indices in the $i$-th block and $Y_{B_i}$ denote the corresponding subsequence. 
% In the forward process, We applies a monotonically increasing noise schedule ($t_1 < t_2 < \dots < t_N$) to the $N$ blocks, i.e., $Y_{B_{<N}}^{t_{<N}} = \{Y_{B_{1}}^{t_{1}}, \dots, Y_{B_{N}}^{t_{N}} \}$. 
% Namely, the earlier blocks in $Y_{B_{<N}}^{t_{<N}}$ are progressively less masked (i.e., more complete), while later blocks remain increasingly masked (i.e., more uncertain).. 
% For the reverse process,~\method~trains a $\theta$-parameterized model to characterize:
% \begin{equation}
%     p_\theta(Y^0 | Y^t)=\prod_{i=1}^{N} p_\theta(Y_{B_{i}}^{0}| Y_{B_{1}}^{t_{1}}, \dots, Y_{B_{i}}^{t_{i}} ). 
% \end{equation}
% Intuitively, 
% the learned model can first finalize the decoding of preceding blocks while simultaneously advancing the denoising of subsequent ones, which effectively enables inter-block parallel decoding. 
Let the index set of the $i$-th block be defined as 
$B_i = \{(i-1)k, \ldots, ik-1\}$, and denote by $Y_{B_i}$ the corresponding token subsequence. 

During the forward diffusion process, we assign progressively increasing noise levels to different blocks according to a monotonic schedule $t_1 < t_2 < \cdots < t_N$. 
Formally, the noise sequence can be represented as 
$
Y^{t_{1:N}} = \{Y_{B_1}^{t_1}, \ldots, Y_{B_N}^{t_N}\}.
$
Under this design, earlier blocks are exposed to lower corruption levels and thus retain more complete information, whereas later blocks remain more heavily masked and uncertain.

For the reverse process, we learn a $\theta$-parameterized model that factorizes the conditional distribution in a block-wise autoregressive manner:
\begin{equation}
    p_\theta(Y^0 \mid Y^{t_{1:N}}) 
    = \prod_{i=1}^{N} 
    p_\theta\!\left(
        Y_{B_i}^0 
        \mid 
        Y_{B_1}^{t_1}, \ldots, Y_{B_i}^{t_i}
    \right).
\end{equation}

This formulation allows the model to progressively refine earlier blocks while concurrently denoising later ones, naturally enabling parallel decoding across blocks without sacrificing temporal consistency.

% \subsection{Training: Asymmetric Distillation}
\subsection{Training: Asymmetric Distillation for Efficient Post-Training}
\label{sec:train}
% 传统的block-difussion dllm在训练会构造两倍的序列长度来进行训练，因为预测当前nosie block需要保证之前的block是干净的，一般会把干净的整条序列拼接在noise序列之前（引用一下 block difussion），这样会导致训练成本增加。受worldvla启发（不同chunk间的动作token互不关注效果更好），或者说预测下一个chunk不需要基于完全干净之前的chunk，我们构造逐级递减的噪声序列，保证序列长度和之前一样。
% 我们探索了两种block difussion vla的训练方式，第一种是从基于训练好的双向dvla用不对称蒸馏的方法训练，第二种是从预训练的自回归vla，vlm用cross entrovy的方式进行训练。（不对称蒸馏在训练步数上会有优势，并且由于对齐双向dvla的特征，会保留未来block的信息0
% 贴公式

% In traditional block-diffusion training for Diffusion LLMs, the input sequence length is often doubled. This is because predicting the current noisy block requires the preceding blocks to be fully clean, and prior work (e.g., Block Diffusion) typically concatenates a clean sequence before the noisy sequence. Such a design substantially increases the training cost.
To train our \method, a straightforward approach is to train it from scratch while maintaining block-wise attention and diffusion-forcing objective. 
% we directly optimize the model using a cross-entropy objective under the block-diffusion formulation. 
The loss function is defined as:
\begin{equation}
\label{eq:bd_loss}
\mathcal{L}_{\text{BD}}
= \mathbb{E}\sum_{i=1}^{N} 
\left[
    -\log p_{\theta}(Y_{B_i}^0 |Y_{B_<i}^{t_<i},c)
\right].
\end{equation}
 
% 为了低成本、高效率训练，我们从开源dvla模型继承权重，将它们adapt成使用block-wise注意力的，并且适配块间并行的fast block dvla
However, motivated by \Cref{sec:motivations}, directly inheriting the decoding nature from open-source bidirectional dVLAs~\citep{chen2025unified,dreamvla25,liang2025discrete} (serving as teacher models) can be a more efficient and lower-cost way.
Specifically, inspired by \citep{wang2026diffusion}, we design an asymmetric distillation in which the \method~(serving as student models) with block-wise attention is forced to align with the output of the teacher model with bidirectional attention, while they share the same architecture and both condition on the blocks with a monotonic noise schedule.
% Notably, this approach is more efficient and lower-cost in training compared to training from scratch.
% This approach provides faster convergence in training steps and preserves future-block information by aligning with bidirectional features.
Thus, the distillation loss is formulated as:
\begin{equation}
\label{eq:d2f_loss}
\mathcal{L}_\text{AD} = \mathbb{E} \left[ \sum_{i=1}^{N} D_{\text{KL}}\left( p_\theta(Y_{B_i}^0 |Y_{B_{<=i}}^{t_{<=i}},c) \Vert p_{\phi^{-}}(Y_{B_{i}}^0 |Y_{B_{<=N}}^{t_{<=N}},c) \right) \right],
\end{equation}
% where $p_{\phi^{-}}$ denotes a pretrained bidirectional dVLA serving as the teacher model with stopped gradients~\citep{song2025ceed} and $\mathcal{D_{\text{KL}}}$ represents a KL divergence.
where $D_{\text{KL}}$ represents the KL divergence aggregated over the mask tokens. 
The distillation is asymmetric in that the teacher $p_{\phi^-}$ predicts for each block $Y_{B_i}^0$ with a global view of all blocks, while the student $p_\theta$ learns to approximate using only a causally restricted view. 
% They share the same model architecture.

Taking the training budget required for training a dVLA from scratch as the reference, \Cref{fig:training_step} shows that asymmetric distillation from finetuned weight ($\mathcal{L}_{\text{AD}}$ in \Cref{eq:d2f_loss}) achieves convergence with only 1/10 steps, which is much more efficient than training with $\mathcal{L}_{\text{BD}}$ on the base of finetuned weight or from scratch. Thus, we adopt asymmetric distillation as the default training objective.

% As for model architecture, the student differs from the teacher solely in attention masks---it uses the block-wise causal attention instead of a bidirectional one. 

% \Cref{fig:training_step} shows that
% Asymmetric Distillation converges with only a tenfold scaling of the distillation steps relative to fine-tuning, whereas training with $\mathcal{L}_{\text{BD}}$ requires nearly half of the full fine-tuning steps to reach convergence.

\subsection{Inference: Pipelined Parallel Decoding}
\label{sec:infer}          
\begin{figure*}[t]  
    \centering
    \includegraphics[width=0.99\linewidth]{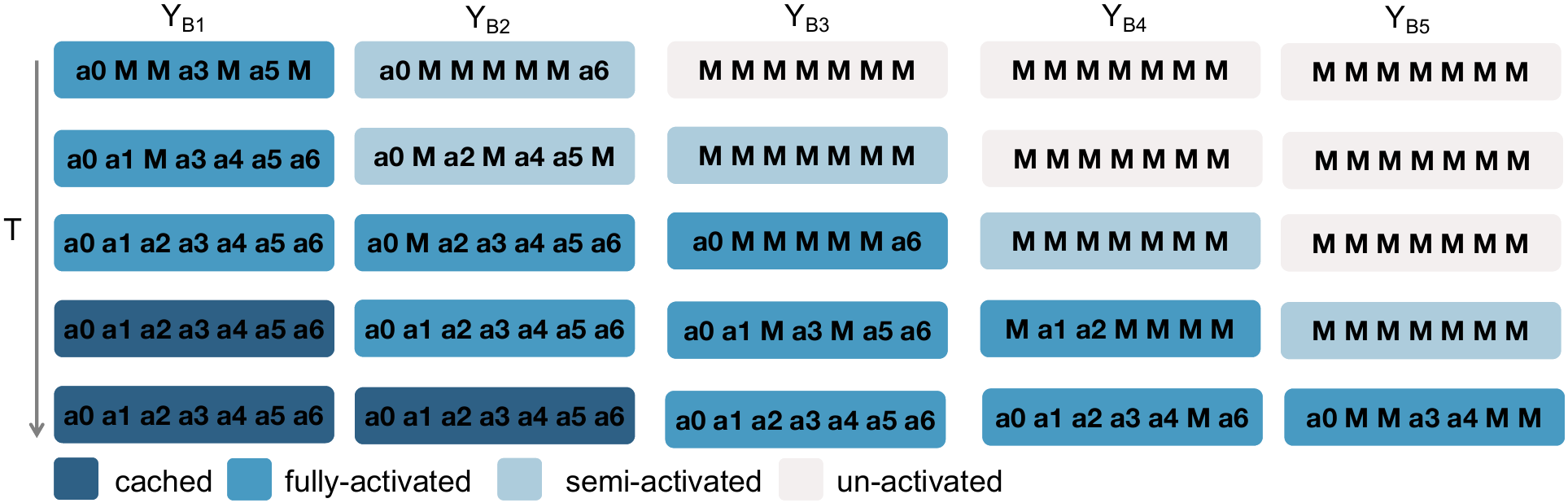} 
    % \caption{
    % Overview of the pipelined parallel decoding strategy in~\method.
    % Instead of processing blocks sequentially, multiple blocks are maintained concurrently in a dynamically growing pipeline.
    % Once the completion ratio of the current tail block surpasses the addition threshold $\tau_{\text{add}} = \tfrac{2}{7}$, a subsequent block is introduced into the pipeline.
    % This newly introduced block starts in a semi-activated mode and becomes fully activated after its preceding block exceeds the activation threshold $\tau_{\text{act}} = \tfrac{4}{7}$.
    % Compared with semi-activated blocks, fully activated ones decode tokens more aggressively by advancing multiple tokens per decoding step.
    % }
    \caption{
    Overview of the pipelined parallel decoding in~\method.
    Blocks are processed concurrently in a dynamically growing pipeline.
    A new block is introduced when the tail block exceeds the addition threshold $\tau_{\text{add}}=\tfrac{2}{7}$, and becomes fully activated after its predecessor surpasses the activation threshold $\tau_{\text{act}}=\tfrac{4}{7}$.
    % Fully activated blocks解码策略更激进，在一次迭代中相比semiactivated blocks会解码更多的action token
    % Fully activated blocks decode multiple tokens per step.
    }

    \label{fig:decode} 
    \vspace{-0.5cm}
\end{figure*}

As shown in~\Cref{fig:decode}, in contrast to traditional block diffusion~\citep{arriola2025block}, which performs parallel decoding only within each block while strictly decoding different blocks in sequence, our method enables parallel prediction across multiple blocks.

Specifically, we distinguish activated blocks (i.e., the block currently being decoded) into two states: \emph{semi-activated} and \emph{fully-activated}. The transition between these states is governed by the completion ratio of the preceding block with respect to the thresholds $\tau_{\text{add}}$ and $\tau_{\text{act}}$. When the completion ratio of the previous block exceeds $\tau_{\text{add}}$, the subsequent block is introduced as a semi-activated block. We adopt the confidence-aware decoding strategy~\citep{wu2025fast} to selectively decode tokens whose prediction confidence exceeds the threshold $\tau_{\text{conf}}$. Once the completion ratio surpasses $\tau_{\text{act}}$, the block transitions to a fully-activated state, in which at least $1/n$ of the remaining tokens are guaranteed to be decoded at each step according to confidence ranking.

This multi-state block-parallel decoding mechanism achieves a favorable trade-off between efficiency and performance. At the same time, it ensures that earlier action tokens are decoded in early iterations, thereby preserving the temporal causality inherent in action execution. A pseudocode summary of our inference is available in the supplementary material.

\section{Experiments}
We conduct comprehensive experiments to evaluate the effectiveness of \method~in both simulated and
real-world robot manipulation tasks. The experiments are designed to answer five core questions:

\noindent
\textbf{(RQ1)} Does our \method~achieve a favorable performance/speed trade-off among all dVLA acceleration paradigms? Furthermore, is \method~consistently effective across diverse dVLA architectures (see \Cref{sec: paradigm})?

\noindent
\textbf{(RQ2)} How does the performance of existing dVLA methods accelerated by \method~compare to SOTA methods (\textit{i.e.}, flow-matching VLAs) on diverse benchmarks and tasks (see \Cref{sec:sota})?

\noindent
\textbf{(RQ3)} Could \method~facilitate the real-world tasks (see \Cref{sec:real})?

\noindent
\textbf{(RQ4)} Is the training of \method~efficient (see \Cref{sec:efficiency})?

\noindent
\textbf{(RQ5)} What empirical insights can guide the selection of hyperparameters for \method~to ensure optimal performance (see \Cref{sec:ablation}) ?

\subsection{Setup}
% \textbf{Base Model.}
% 我们选择UD-VLA作为Unified dVLA的代表，以及选择dream-vla，discrete difussion vla作为dVLA的代表。对于UD-VLA，我们遵循原工作的设定在calvin ABCD-D做了实验，设置action chunk=10,将第三视角图像离散化25*25=625，将未来腕部图像离散化为10*10=100个token，我们在蒸馏步数设置为原udvla微调步数的1/8，也就是3k步，batch大小设置为12，由于输出序列较长，我们将block siez 设置为32的倍数。其它的训练设定与udvla一致对于。对于dreamvla，我们参考原来的模型对libero任务的设定，我们设置action chunk为8，对于simpler 任务，我们设置为5,对于libero和simpler任务我们均将训练步数设置4k步，大致为微调步数的1/5，对于discrete difussion vla，我们将对libero的蒸馏步数设置为4k步，大致为原微调步数的1/8，二者的batch大小设置为8，我们将block_size设置为7的倍数，这样可以保证block不会切分同一时间步的动作，其它的训练参数均与原模型一致。
% 我们采用搭载lora方法进行不对称蒸馏，我们将rank设置为32,在蒸馏过程中，我们将教师模型视为不挂在lora模块的经过微调后的模型，在计算教师logit时我们不激活lora分支，计算学生模型时我们激活lora分支，这样我们最大保存了原有的dvla backbione模型语言视觉理解以及动作推理先验，lora模块仅仅学习注意力方式的迁移

\noindent
\textbf{Models.} We select Dream-VLA and DD-VLA as representatives of dVLA models, and UD-VLA as the representative of unified dVLA models.
For Dream-VLA, 
% we follow the original settings used for LIBERO tasks, setting the action chunk size to 8, and to 5 for SIMPLER tasks. 
% For both Dream-VLA and DD-VLA, the batch size is set to 8. 
% For both LIBERO and SIMPLER, 
we perform distillation for 4k steps, corresponding to approximately $1/5$ of the original fine-tuning budget.
For DD-VLA, we set the distillation steps to 4k, which is approximately $1/8$ of the original fine-tuning steps. 
We set the block size to be 7, as analyzed in \Cref{sec:ablation}.
% a multiple of 7 to ensure that blocks do not split action tokens belonging to the same time step. 
For UD-VLA, 
% we follow the official setting on CALVIN. 
% Specifically, we set the action chunk size to 10, discretize third-person observations into $25\times25=625$ tokens, and discretize wrist-view images into $10\times10=100$ tokens. 
we perform distillation for 3k steps, corresponding to approximately $1/8$ of the original UD-VLA fine-tuning steps, with a batch size of 12. Due to the relatively long output sequences (625 tokens) in UD-VLA, we set the block size to be a multiple of 32. All remaining training hyperparameters follow the original model configurations.

\noindent
\textbf{Benchmarks.}
We conduct extensive simulated experiments on three popular benchmarks (CALVIN~\citep{mees2022calvin}, LIBERO~\citep{liu2023libero}, and SimplerEnv~\citep{li24simpler}) to provide comprehensive results.
Detailed introduction of these benchmarks is available in the supplementary material.

\subsection{Paradigm Comparison (RQ1)}
\label{sec: paradigm}

% \begin{table*}[t]
% % \setlength{\tabcolsep}{4pt}
% % \renewcommand{\arraystretch}{0.9}
% \centering
% \small
% \caption{Comparison with various base models in terms of inference speed, task-wise success rates (SR), average SR, and execution frequency.}
% \label{tab:libero_ablation}
% \resizebox{\textwidth}{!}{%
% \begin{tabular}{@{} l c c c c c c c @{}}
% \toprule
% \multirow{2}{*}{Decoding Method} 
% & \multirow{2}{*}{Speed (Tokens/s)} 
% & \multicolumn{5}{c}{Success Rate (\%)} 
% & \multirow{2}{*}{Freq. (Hz)} \\
% \cmidrule(lr){3-7}
% & & Spatial & Goal & Object & Long & Avg. & \\
% \midrule
% Dream-VLA~\citep{yedreamVLA}
% & 98.8
% & 0.902 & 0.920 & 0.880 & 0.720 & 0.856
% &  \\

% \hspace{3mm} + Fast-dLLM  
% & 183.2
% & 0.884 &  0.894 &0.834 &  0.702 &  0.828  
% &  \\

% \hspace{3mm} + Block Diffusion  
% & 181.7 
% & 0.918 &0.904 & 0.886 &0.722  &  0.858
% &  \\

% \rowcolor[gray]{0.9} \hspace{3mm} + ours 
% & \textbf{313.1} 
% & 0.912 & 0.92 & 0.902 & 0.746 & 0.870 
% &  \\
% \midrule
% Discrete Difussion VLA~\citep{liang2025discrete}
% & 152.1
% & 0.972 & 0.986 & 0.974 & 0.920 & 0.963
% & - \\

% \hspace{3mm} + Fast-dLLM  
% & 312.5
% & 0.940 & 0.952 & 0.948 & 0.898 & 0.935
% & - \\

% \hspace{3mm} + Block Diffusion  
% & 322.1
% & 0.976 & 0.986 & 0.972 & 0.932 & 0.967
% & - \\

% \rowcolor[gray]{0.9} \hspace{3mm} + ours 
% & \textbf{402.7} 
% & 0.970 & 0.988 & 0.976 & 0.928 & 0.966
% & - \\

% \bottomrule
% \end{tabular}
% }
% % \vspace{-0.2cm}
% \end{table*}

\begin{table*}[t]
\setlength{\tabcolsep}{5pt}
\renewcommand{\arraystretch}{1}
\centering
\small
\caption{Comparison between various acceleration strategies on two base models in terms of task-wise success rates (SR) and inference speed on LIBERO.}
\label{tab:libero_ablation}
\resizebox{0.9\textwidth}{!}{%
\begin{tabular}{@{} l c c c c c c @{}}
\toprule
\multirow{2}{*}{Decoding Method} 
& \multicolumn{5}{c}{Success Rate $\uparrow$ } 
& \multirow{1}{*}{Speed $\uparrow$ } \\
\cmidrule(lr){2-6}
& Spatial & Goal & Object & Long & Avg. &(Tokens/s)  \\
\midrule

Dream-VLA~\citep{yedreamVLA}
& 0.902 & 0.920 & 0.880 & 0.720 & 0.856
& 98.8 \tiny{(\textit{×1.0})} \\

\hspace{3mm} + Fast-dLLM  
& 0.884 & 0.894 & 0.834 & 0.702 & 0.828
& 183.2 \tiny{(\textit{×1.9})} \\

\hspace{3mm} + Block Diffusion  
& 0.918 & 0.904 & 0.886 & 0.722 & 0.858
& 181.7 \tiny{(\textit{×1.8})} \\

\rowcolor[gray]{0.9} \hspace{3mm} + \method~(ours) 
& 0.912 & 0.920 & 0.902 & 0.746 & 0.870
& \textbf{313.1} \tiny{(\textit{×3.2})} \\

\midrule

Discrete Diffusion VLA~\citep{liang2025discrete}
& 0.972 & 0.986 & 0.974 & 0.920 & 0.963
& 152.1 \tiny{(\textit{×1.5})} \\

\hspace{3mm} + Fast-dLLM  
& 0.940 & 0.952 & 0.948 & 0.898 & 0.935
& 312.5 \tiny{(\textit{×3.2})} \\

\hspace{3mm} + Block Diffusion  
& 0.976 & 0.986 & 0.972 & 0.932 & 0.967
& 322.1 \tiny{(\textit{×3.3})} \\

\rowcolor[gray]{0.9} \hspace{3mm} + \method~(ours)
& 0.970 & 0.988 & 0.976 & 0.928 & 0.966
& \textbf{402.7} \tiny{(\textit{×4.1})} \\

\bottomrule
\end{tabular}
}
\end{table*}

\noindent
\textbf{\method~achieves obvious acceleration with competitive performance.}
%相对backbone加速了多少
% As shown in \Cref{tab:libero_ablation}, DreamVLA and Discrete Difussion VLA在保证加速比2.92倍以上的同时，与base mdoel相比，我们的性能还取得了些许提升，原因可能是我们的fast block difussion以及block difussion的方式，保证了推理动作在时间维度上的因果性，与全双向的dvla相比，我们的block-wise的注意力设计更符合逐action chunk的模式，更容易对动作分布进行拟合。
\Cref{tab:libero_ablation} shows that Fast-dLLM~\citep{wu2025fast} realizes 2$\times$ speedup with an obvious performance drop.
This proves that directly reusing the KV cache under fully bidirectional attention introduces biased keys and values in the attention states that lead to performance degradation (see \Cref{fig:attn_full}).
Block Diffusion~\citep{arriola2025block} decodes blocks strictly in sequence, thus also obtaining a limited acceleration.
In contrast, our \method~built on Dream-VLA and DD-VLA both achieve speedups up to 4.1$\times$, while slightly improving performance compared to their base models. 
We attribute the efficiency gain to the discrete diffusion forcing denoising, which effectively reuses the KV cache and offers inter-block parallelism, increasing throughput per iteration step. 
% 有效利用kv缓存，
Meanwhile, our block-wise attention preserves temporal causality in action prediction, leading to more stable optimization during training.
% Compared to Block Diffusion~\citep{arriola2025block}, which decodes blocks strictly in sequence, our method enables parallel decoding of multiple action blocks. 
In addition, the results further demonstrate the effectiveness of our acceleration strategy across different methods on the same dataset.

% \subsection{Extending \method~to Unified dVLAs}

\noindent
\textbf{\method~naturally generalizes to unified dVLA architectures.}
As shown in \Cref{tab:calvin_ablation}, \method~achieves a $2.8\times$ inference speedup over UD-VLA on the long-horizon CALVIN ABCD-D benchmark, while maintaining superior performance.
This result demonstrates that our method can be seamlessly extended to unified dVLA frameworks that generate visual foresights together with actions to serve as a process of chain of thought, highlighting its adaptability in accelerating multimodal generation and action prediction.
% adaptability to visual generation tasks and high-level planning scenarios.

% In contrast, Fast-dLLM~\citep{wu2025fast}, which directly applies fully bidirectional caching, introduces inconsistencies during inference and consequently degrades performance. 

% By combining activation thresholds with a progressively decaying noise schedule, Fast-dVLA achieves a higher acceleration ratio while preserving performance consistency.

% \begin{table}[t]
% \setlength{\tabcolsep}{5pt}
% \renewcommand{\arraystretch}{0.9}
% \centering
% \footnotesize
% \caption{Comparison with various decoding methods in terms of tasks completed in a row, average trajectory length, and inference speed on UD-VLA.}
% \label{tab:calvin_ablation}
% \resizebox{\textwidth}{!}{
% \begin{tabular}{@{} l c c c c c c c c @{}}
% \toprule
% Decoding Method 
% & Splits
% & \multicolumn{5}{c}{Tasks Completed in a Row (\%)} 
% & Avg. 
% & Speed  \\
% \cmidrule(lr){3-7}
% & & 1/5 & 2/5 & 3/5 & 4/5 & 5/5 &Len. &(Tokens/s) \\
% \midrule

% UD-VLA  
% & ABCD$\rightarrow$D 
% & 0.992 & 0.968 & 0.936 & 0.904 & 0.840 
% & 4.64 
% & 67.3 \\

% + Fast-dLLM
% & ABCD$\rightarrow$D 
% & 0.972 & 0.920 & 0.858 & 0.808 & 0.762 
% & 4.32 
% & 132.5 \\

% + Block Diffusion  
% & ABCD$\rightarrow$D 
% & 0.988 & 0.944 & 0.894 & 0.862 & 0.804 
% & 4.50 
% & 129.5 \\

% \rowcolor[gray]{0.9} + ours
% & ABCD$\rightarrow$D 
% & 0.984 & 0.952 & 0.922 & 0.870 & 0.812 
% & 4.54 
% & \textbf{186.7} \\

% \bottomrule
% \end{tabular}
% }
% \vspace{-0.2cm}
% \end{table}
\begin{table}[t]
\setlength{\tabcolsep}{5pt}
\renewcommand{\arraystretch}{1}
\centering
\small
\caption{Comparison between various acceleration strategies on UD-VLA in terms of tasks completed in a row, average sentence length, and inference speed on CALVIN.}
\label{tab:calvin_ablation}
\resizebox{\textwidth}{!}{
\begin{tabular}{@{} l c c c c c c c c @{}}
\toprule
\multirow{2}{*}{Decoding Method} 
& \multirow{2}{*}{ Task} 
& \multicolumn{5}{c}{Tasks Completed in a Row $\uparrow$} 
& Avg. 
& Speed $\uparrow$ \\
\cmidrule(lr){3-7}
& & 1/5 & 2/5 & 3/5 & 4/5 & 5/5 &Len. $\uparrow$ & (Tokens/s) \\
\midrule

UD-VLA  
& ABCD$\rightarrow$D 
& 0.992 & 0.968 & 0.936 & 0.904 & 0.840 
& \textbf{4.64 }
& 67.3 \tiny{(\textit{×1.0})} \\

+ Fast-dLLM
& ABCD$\rightarrow$D 
& 0.972 & 0.920 & 0.858 & 0.808 & 0.762 
& 4.32 
& 132.5 \tiny{(\textit{×2.0})} \\

+ Block Diffusion  
& ABCD$\rightarrow$D 
& 0.988 & 0.944 & 0.894 & 0.862 & 0.804 
& 4.50 
& 129.5 \tiny{(\textit{×1.9})} \\

\rowcolor[gray]{0.9} + \method~(ours)
& ABCD$\rightarrow$D 
& 0.984 & 0.952 & 0.922 & 0.870 & 0.812 
& 4.54 
& \textbf{186.7} \tiny{(\textit{×2.8})} \\

\bottomrule
\end{tabular}
}
\vspace{-0.2cm}
\end{table}

\subsection{Comparison with SOTA (RQ2)}
\label{sec:sota}

\noindent
\textbf{Accelerating UD-VLA on CALVIN.} 
%与world-modelling VLAs that同时生成未来图像和动作的相比，our accelerated unified dVLA 仍然保持了SOTA的性能。FAST-dVLA继承了UD-VLA强大的表达能力和统一的多模态空间的优越性，并通过加速在一定程度上解决了future image token所需要的长序列的输出时间。使得unified dvla成为更有竞争力的world-modelling VLA方案。
\Cref{tab:calvin_results} shows that compared with world-modeling VLA~\citep{zhang2025upvla} that jointly generate future images and actions, our~\method~applied on UD-VLA inherits the strong representational capacity and the advantages of a unified multimodal latent space, while mitigating the long decoding latency induced by future image tokens through our acceleration technique. 
These results demonstrate that the unified dVLA paradigm can serve as a practical and viable solution for VLAs with a world-modeling process.

\noindent
\textbf{Accelerating Dream-VLA on SimplerEnv.} 
% 在更符合真实世界测试环境的Simpler上，我们的性能与在libero以及calvin上的结果具有一致性，相对于传统的自回归范式openvla以及π0+FAST，我们并行解码的范式可以大幅度缩短解码时间，受益于离散difussion vla优异的动作推理性能模态对其能力，我们的方法在保证加速效果先进的同时，性能也领先连续difussion推理范式的GR00T-N1 以及pi0
On the SimplerEnv, which more closely reflects real-world robotic evaluation settings with high visual fidelity, our results are consistent with those observed on CALVIN. 
\Cref{tab:windowx} shows that our \method~achieves the highest decoding speeds among all VLAs with discrete outputs, including AR paradigms (OpenVLA and $\pi_0$-FAST), vanilla dVLA paradigms (DD-VLA and LLaDA-VLA), and block diffusion paradigms (Dream-VLA with Fast-dLLM or Block Diffusion).
This improvement stems from the effective combination of KV caching and inner/inter-block parallelism.

In terms of task success rates, benefiting from the superior cross-modal alignment of dVLAs, our \method~outperforms continuous flow-matching approaches such as GR00T-N1 and $\pi_0$.
Furthermore, leveraging the sequential action representation induced by our method and the large-scale robot pretraining of Dream-VLA, our \method~also surpasses existing dVLA methods.

For the comprehensive comparison on LIBERO, please refer to the supplementary materials.

\begin{table}[t]
    \footnotesize 
    \setlength{\tabcolsep}{5pt}
    \centering
    \caption{Comprehensive evaluation of long-horizon manipulation on the CALVIN benchmark. UniVLA$^{*}$ denotes the variant without historical frames for fair comparison.}
    \vspace{-0.3cm}
    \resizebox{0.95\textwidth}{!}{
    \begin{tabular}{l c c c c c c c}
        \toprule
        \multirow{2}{*}{\textbf{Method}}   & \multirow{2}{*}{\textbf{Task}} & \multicolumn{5}{c}{\textbf{Tasks Completed in a Row}} & \multirow{2}{*}{\textbf{Avg. Len. $\uparrow$}} \\
        \cmidrule(lr){3-7}
        &  & 1/5 & 2/5 & 3/5 & 4/5 & 5/5 & \\
        \midrule
        % MCIL~\citep{lynch2020language} & ABCD$\rightarrow$D &0.373 & 0.027 & 0.002 & 0.000 & 0.000 & 0.40\\
        RT-1~\citep{brohan2022rt}  & ABCD$\rightarrow$D & 0.844 & 0.617 & 0.438 & 0.323 & 0.227 & 2.45 \\
        % Robo-Flamingo~\citep{li2024vision} & ABCD$\rightarrow$D & 0.964 & 0.896 & 0.824 & 0.740 & 0.660 & 4.09 \\
        LLaDA-VLA~\citep{wen2025llada} & ABCD$\rightarrow$D & 0.956 & 0.878 & 0.795 & 0.739 & 0.645 & 4.01 \\
        Deer~\citep{yue2024deer} & ABCD$\rightarrow$D & 0.982 & 0.902 & 0.821 & 0.759 & 0.670 & 4.13 \\
        GR-1~\citep{wu2023unleashing}  & ABCD$\rightarrow$D & 0.949 & 0.896 & 0.844 & 0.789 & 0.731 & 4.21 \\
        ReconVLA~\citep{song2025reconvla}  & ABCD$\rightarrow$D & 0.980 & 0.900 & 0.845 & 0.785 & 0.705 &  4.23 \\
        UniVLA$^{*}$~\citep{univla}  & ABCD$\rightarrow$D & 0.948 & 0.906 & 0.862 & 0.834 & 0.690 & 4.26 \\
        MODE~\citep{reussefficient}  & ABCD$\rightarrow$D & 0.971 & 0.925 & 0.879 & 0.835 & 0.779 & 4.39 \\
        UP-VLA~\citep{zhang2025upvla}  & ABCD$\rightarrow$D & 0.962 & 0.921 & 0.879 & 0.842 & 0.812 & 4.42 \\
        MDT~\citep{reuss2024multimodal}  & ABCD$\rightarrow$D & 0.986 & 0.958 & 0.916 & 0.862 & 0.801 & 4.52 \\
        % RoboVLMs~\citep{li2024towards} & ABCD$\rightarrow$D & 0.967 & 0.930 & 0.899 & 0.865 & 0.826 & 4.49 \\
        \rowcolor[gray]{0.9} \textbf{UD-VLA + \method~(ours)}  & ABCD$\rightarrow$D  & 0.984 &  0.952 & 0.922 &  0.870 & 0.812  & \textbf{4.54}\\
        \bottomrule
    \end{tabular}
    }
    \label{tab:calvin_results}
    % \vspace{-0.5cm}

\end{table}

\begin{table*}[t]
\centering
\caption{Evaluation on WidowX Robot tasks in SimplerEnv. We report the Grasping Success Rate (Grasp) and Task Success Rate (Success) in percentages (\%). Besides, we further report the decode speed (Speed) of all VLAs with discrete output, which is calculated in tokens per second.}
% \vspace{-0.3cm}
\label{tab:windowx}
\resizebox{0.95\textwidth}{!}{%
\begin{tabular}{lcccccccccc}
\toprule
\multirow{2}{*}{\textbf{Method}} 
& \multicolumn{2}{c}{\textbf{Spoon on Towel}} 
& \multicolumn{2}{c}{\textbf{Carrot on Plate}} 
& \multicolumn{2}{c}{\textbf{Stack Green Block}} 
& \multicolumn{2}{c}{\textbf{Eggplant in Basket}} 
& \multicolumn{2}{c}{\textbf{Average}} \\
\cmidrule(lr){2-3} \cmidrule(lr){4-5} \cmidrule(lr){6-7} \cmidrule(lr){8-9} \cmidrule(lr){10-11}
 & Grasp & Success & Grasp & Success & Grasp & Success & Grasp & Success & Success & Speed \\
\midrule

% Octo-Base~\citep{octo_2023} 
% & 50.0 & 33.0 
% & 50.0 & 25.0 
% & 29.2 & 0.0 
% & 40.0 & 23.3 
% & 20.3 & - \\

% RT-1-X~\citep{brohan2022rt} 
% & 4.2 & 0.0 
% & 16.7 & 0.0 
% & 0.0 & 0.0 
% & 3.3 & 0.0 
% & 0.0 & - \\

% OpenVLA~\citep{kim2024openvla} 
% & 4.1 & 0.0 
% & 33.0 & 0.0 
% & 12.5 & 0.0 
% & 8.3 & 4.1 
% & 1.0 & - \\

RoboVLM~\citep{li2024towards} 
& 37.5 & 20.8 
& 33.3 & 25.0 
& 8.3 & 8.3 
& 0.0 & 0.0 
& 13.5 & - \\

SpatialVLA~\citep{qu2025spatialvla} 
& 20.8 & 16.7 
& 29.2 & 25.0 
& 62.5 & 29.2 
& \textbf{100.0} & \textbf{100.0} 
& 42.7 & - \\

OpenVLA-OFT~\citep{kim2025fine} 
& 50.0 & 12.5 
& 41.7 & 4.2 
& 70.8 & 20.8 
& 91.7 & 37.5 
& 18.8 & -\\

$\pi_0$~\citep{Pi0} 
& 45.8 & 29.1 
& 25.0 & 0.0 
& 50.0 & 16.6 
& 91.6 & 62.5 
& 27.1 & - \\

$\pi_0$-FAST~\citep{pertsch2025fast} 
& 62.5 & 29.1 
& 58.5 & 21.9 
& 54.0 & 10.8 
& 83.3 & 66.6 
& 32.1 & 107.5 \\

GR00T-N1~\citep{gr00t} 
& \textbf{83.3} & \textbf{62.5} 
& 54.2 & 45.8 
& 70.8 & 16.7 
& 41.7 & 20.8 
& 36.5 & - \\

DDVLA~\citep{liang2025discrete} 
& 70.8 & 29.2 
& 58.3 & 29.2 
& 62.5 & 20.8 
& 91.7 & 70.8 
& 37.5 & 152.8 \\
LLaDA-VLA~\citep{wen2025llada} 
& - & 56.9 
& - & \textbf{76.3} 
& - & 30.6 
& - & 58.3 
& 55.5 & 160.0 \\

Dream-VLA~\citep{yedreamVLA} 
& 79.2 & 45.8 
& 62.5 & 45.8 
& \textbf{83.3} & 25.0 
& \textbf{100.0} & 87.5 
& 51.0 & 100.1 \\

\hspace{3mm} + Fast-dLLM~\citep{wu2025fast}
& 70.8 & 41.7 
& 54.2 & 37.5 
& 70.8 & 20.8 
& 83.3 & 66.6 
& 41.7 & 214.2 \\

\hspace{3mm} + Block Diffusion~\citep{arriola2025block}
& \textbf{83.3} & 54.1 
& \textbf{66.7} & 45.8 
& \textbf{83.3} & 29.1 
& 95.8 & 91.6 
& 55.2 & 226.4 \\

\rowcolor[gray]{0.9} \hspace{3mm} + \method~(ours) 
& \textbf{83.3} & 54.1 
& 62.5 & 54.1 
& \textbf{83.3} & \textbf{37.5} 
& \textbf{100.0} & 91.6 
& \textbf{59.3} & \textbf{366.4} \\

\bottomrule
\end{tabular}%
}
\vspace{-0.1cm}
\end{table*}

% Acceleration on VLA models with unified multimodal models (UMMs) architectures.
% As shown in \Cref{tab:calvin_results} and  \Cref{tab:calvin_ablation},
% Fast dvla在长序列任务calvinABCD-D,在保证推理速度想读与udvla加速2.78倍的同时，保持性能依旧出于sota水平，我们的加速方法不会损害在长序列任务推理以及执行的性能，相对于fast-dllm的方法，直接应用全双向的缓存，导致不一致的推理结果，导致性能下降，对于Block Diffusion顺序解码block的方式，我们的方法并行解码多个action 块，并拖过激活阈值以及噪声递减调度的得到了更高的加速比，并保证性能一致

% \subsection{Comparison with Continuous Diffusion VLAs}
%讨论理论上的优势，比较速度（可以简单的用latency，也就是推理速度来衡量）和性能

% \subsection{Paradigm Comparison}
% Discrete Autoregressive VLA (OpenVLA) & Discrete Diffusion VLA (DreamVLA) & Block Diffusion VLA & Our Diffusion Forcing VLA
%首先需要画一个图解释四种vla的区别，这个图下面会有个表，表里的数据包括一个完整的sequence输出需要多少次forward，每次forward需要多长时间，从而带来的总的速度。
\subsection{Real-world Experiments (RQ3)}
\label{sec:real}
\begin{figure*}[t]  
    \centering
    \includegraphics[width=0.9\linewidth]{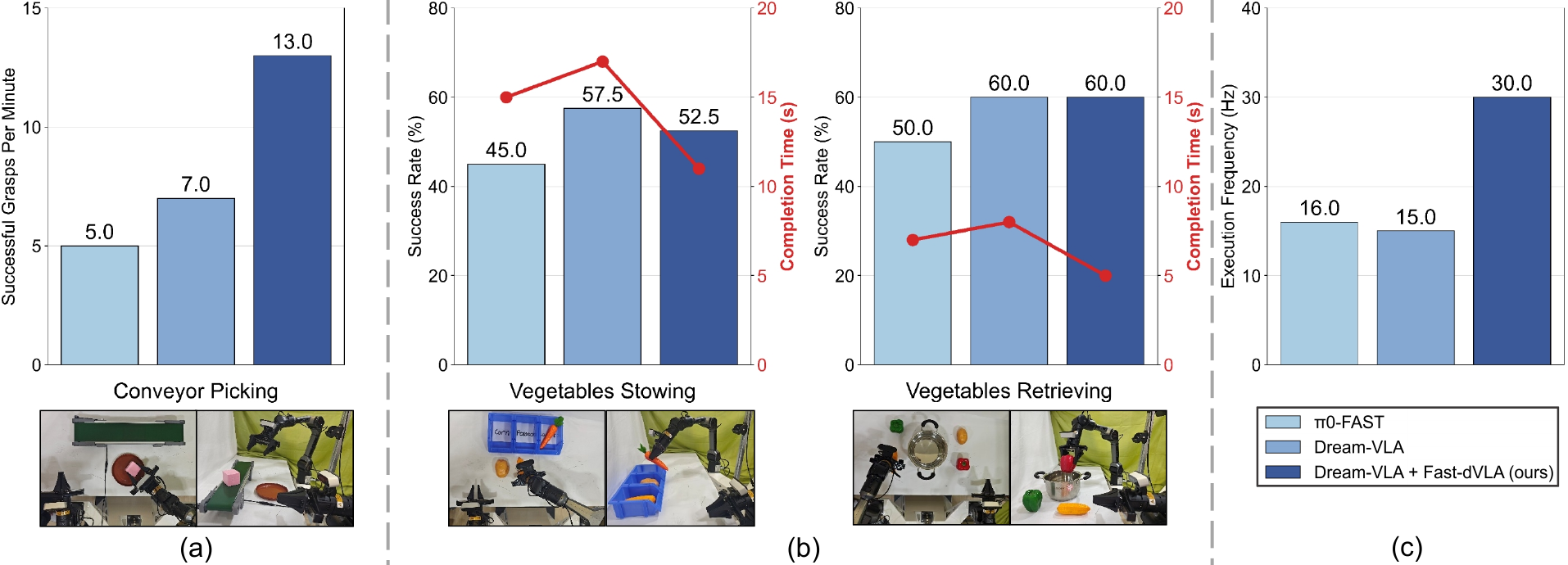} 
    \caption{Real-world experiment results. 
    % We evaluate our Fast-dVLA and prior SOTA VLAs on 3 representative tasks. 
    We report (a) successful grasps per minute ($\uparrow$); (b) success rates ($\uparrow$) and completion times ($\downarrow$); (c) execution frequency ($\uparrow$).}
    % \vspace{-0.5cm}
    \label{fig:realexp} 
\end{figure*}
% 我们在bimanual AgileX上进行我们的真机实验。该机器人每个臂具有六个自由度的关节和一个夹爪，它同时具有一个主相机在中间高处视角，还有手腕处各自一个摄像机。我们一共设置了三个任务，如图所示，分别是在流水线上夹起方块、收纳蔬菜和根据指令捡起蔬菜。对于每个任务，我们均收集一百条数据进行训练。对于评估，我们对每个任务进行了40次评估，记录他们的成功率。特别的，对于流水线夹取任务，我们采用每分钟成功夹取的方块数作为评估指标。此外，我们还记录了三种method在真机上的Execution Frequency
%我们选取了cite{pi0-FASTpertsch2025fast}，Dream-VLA两组具有代表性的模型和我们的方法进行比较：前者是目前离散自回归范式里的sota模型，后者则是离散扩散的代表模型，也是我们的基线。

\noindent
\textbf{Setup.}
Real-world experiments were conducted on a bimanual AgileX platform, where each 6-DOF arm is equipped with a gripper. The sensory suite includes a high-mounted overhead camera providing a global perspective and two wrist-mounted cameras for localized views. 

\noindent
\textbf{Task setting.}
We designed three distinct tasks, as illustrated in \Cref{fig:realexp}: (1) \textit{Conveyor Picking}, which involves picking blocks from a moving conveyor belt and placing them into a tray. (2) \textit{Vegetables Stowing}, which requires sorting vegetables based on their text labels in a container. (3) \textit{Vegetables Retrieving}, which involves grasping a target vegetable and placing it into a pot according to specific language instructions. 
For each task, we collected 100 expert demonstrations for training. 
For evaluation, we conducted 40 trials per task, recording the success rate and the average completion time. 
Specifically, for the conveyor belt task, we utilized the number of successful grasps per minute as the primary evaluation metric to quantify performance. 
In addition, we recorded the execution frequency on the real-world robot platform to quantify the real-time performance. 
% In addition, we measured the execution frequency on the real-world robot.

\noindent
\textbf{Results.}
We evaluate our method against two representative models: $\pi_{0}$-FAST~\citep{pertsch2025fast} (the SOTA AR VLA) and Dream-VLA~\citep{yedreamVLA}, a representative dVLA that acts as our base model. 
\Cref{fig:realexp} shows our model demonstrates robust performance across all three tasks.  Notably, the conveyor belt picking task requires both precise grasping and real-time responsiveness. 
Our method achieves nearly double the efficiency of previous approaches, closely aligned with the practical demands of industrial sorting systems. 
In the remaining two tasks that require semantic understanding, our model maintains competitive performance with only a marginal reduction in success rate relative to the baseline, while further shortening the required completion time. 
Crucially, our system maintains a consistent execution frequency of 30 Hz across all tasks, satisfying the practical demand of real-time control that other approaches fail to meet. 
These results underscore our model's efficient execution and precise instruction-following capabilities.
%如图xx所示\Cref{fig:realexp}，我们的模型在三个任务中均取得不错的表现，特别是传送带夹取这种既需要夹取精度又需要速度的真实任务，我们的方法取得几乎两倍于先前方法的效率。在其余两个任务，我们的模型相对基线的成功率只有轻微的损失，但是任务完成速度却得到进一步的提升。这个结果和我们在仿真benchmark中得到的结论一致，表现了我们模型在真实世界中能胜任既需要时效性又需要语言跟随的任务。我们的Execution Frequency达到了 30Hz，

\subsection{Training Efficiency (RQ4)}
\label{sec:efficiency}
\begin{wrapfigure}{r}{0.5\textwidth}
\centering
\vspace{-0.6cm}
\includegraphics[width=0.99\linewidth]{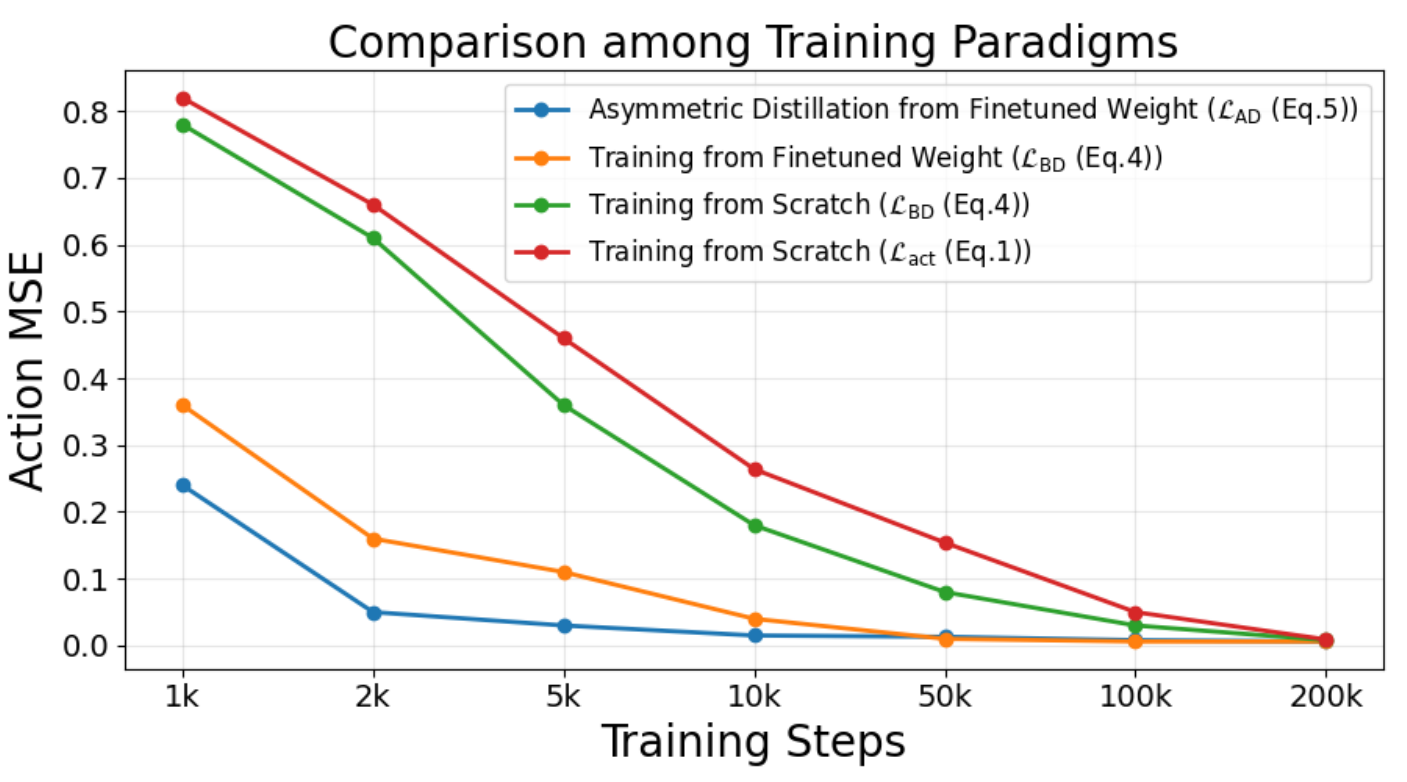}
% \vspace{-0.2cm}
\caption{
Action Mean Squared Error (MSE) of the dVLA at varying training steps on LIBERO. The MSE of our asymmetric distillation exhibits the fastest decline, indicating the most rapid convergence speed.
}
\label{fig:training_step}
\vspace{-0.4cm}
\end{wrapfigure}

To evaluate the training efficiency of our \method, we compare four training strategies based on Dream-VLA on LIBERO:
(1)~\textit{Asymmetric Distillation from Finetuned Weight ($\mathcal{L}_\textnormal{AD}$).} This approach distills our \method~from the task-specific finetuned weight using $\mathcal{L}_\textnormal{AD}$.
(2)~\textit{Training from Finetuned Weight ($\mathcal{L}_\textnormal{BD}$).} This approach trains \method~from the finetuned weight with $\mathcal{L}_\textnormal{BD}$.
(3)~\textit{Training From Scratch ($\mathcal{L}_\textnormal{BD}$).} This approach trains our \method~from the pretrained dVLA.
(4)~\textit{Training From Scratch ($\mathcal{L}_\textnormal{act}$).} 
This approach finetunes a normal dVLA on the specific tasks to serve as the baseline for comparison.

\Cref{fig:training_step} shows that our asymmetric distillation demonstrates significantly superior training efficiency. Notably, the distillation strategy (blue line) requires only 2,000 training steps to converge, which is 5$\times$ faster than continuing to train from the finetuned weight (orange line) and approximately 1/10 of the steps needed for training from scratch (green line).
% These results underscore the efficiency of our proposed distillation strategy. 
From another perspective, our asymmetric distillation offers a cost-efficient pathway to accelerate existing dVLA models to real-time performance required for practical applications.
Besides, all strategies for training our \method~converge faster than finetuning a normal dVLA (red line), denoting that our architecture is more efficient.

\subsection{Ablation Studies (RQ5)}
\label{sec:ablation}
% \noindent

\noindent
\textbf{Block size aligned with action dimensionality brings better performance.}
We validate the importance of choosing the multiples of the action dimensionality as the block size.
To ensure fairness, the results are averaged from several values between the sizes of 7 (action token numbers in one step) and 14.
Specifically, we find that this choice better maintains success rates and speedup, demonstrating its coherence with the action architecture that better keeps the intrinsic temporal dependencies of the action tokens (see~\Cref{tab:blocksize_libero}).

% For block-size selection, we use a block size that is a multiple of 32 for Unified dVLA models that incorporate visual Chain-of-Thought and long-sequence outputs. As shown in \Cref{tab:blocksize_libero}, we perform an ablation sweep over block size to identify the best trade-off between acceleration and performance. 
% For Dream-VLA with a bin tokenizer (used in~\citep{kim2024openvla}), we find that choosing block sizes that are multiples of 7 (matching the dimensionality of action) yields better speed and performance trade-offs than using multiples of 8.

% \begin{wrapfigure}{r}{0.5\textwidth} 
%     % \vspace{-0.8cm}
%     \centering
%     \includegraphics[width=0.99\linewidth]{figure/confidence.pdf} 
%     \caption{
%     % \textbf{example.} 
%     % 加速效果和performance关于置信度的消融在UD-VLA上的消融，我们发现当confidence=0.5时，性能趋于稳定，并且此时加速比达到2.78\time,
%     Ablation study on the confidence threshold~$\tau_{\text{conf}}$ of \method~based on UD-VLA.
%     % shows the trade-off between acceleration and performance.
%     % We choose a confidence threshold of 0.5 for the trade-off.
%     % , at which performance stabilizes while achieving a speedup of $2.78\times$.
%     }
%     \label{fig:abl_conf} 
%     % \vspace{-1.6cm}
% \end{wrapfigure}
\begin{figure}{t}
    % \vspace{-0.8cm}
    \centering
    \includegraphics[width=0.8\linewidth]{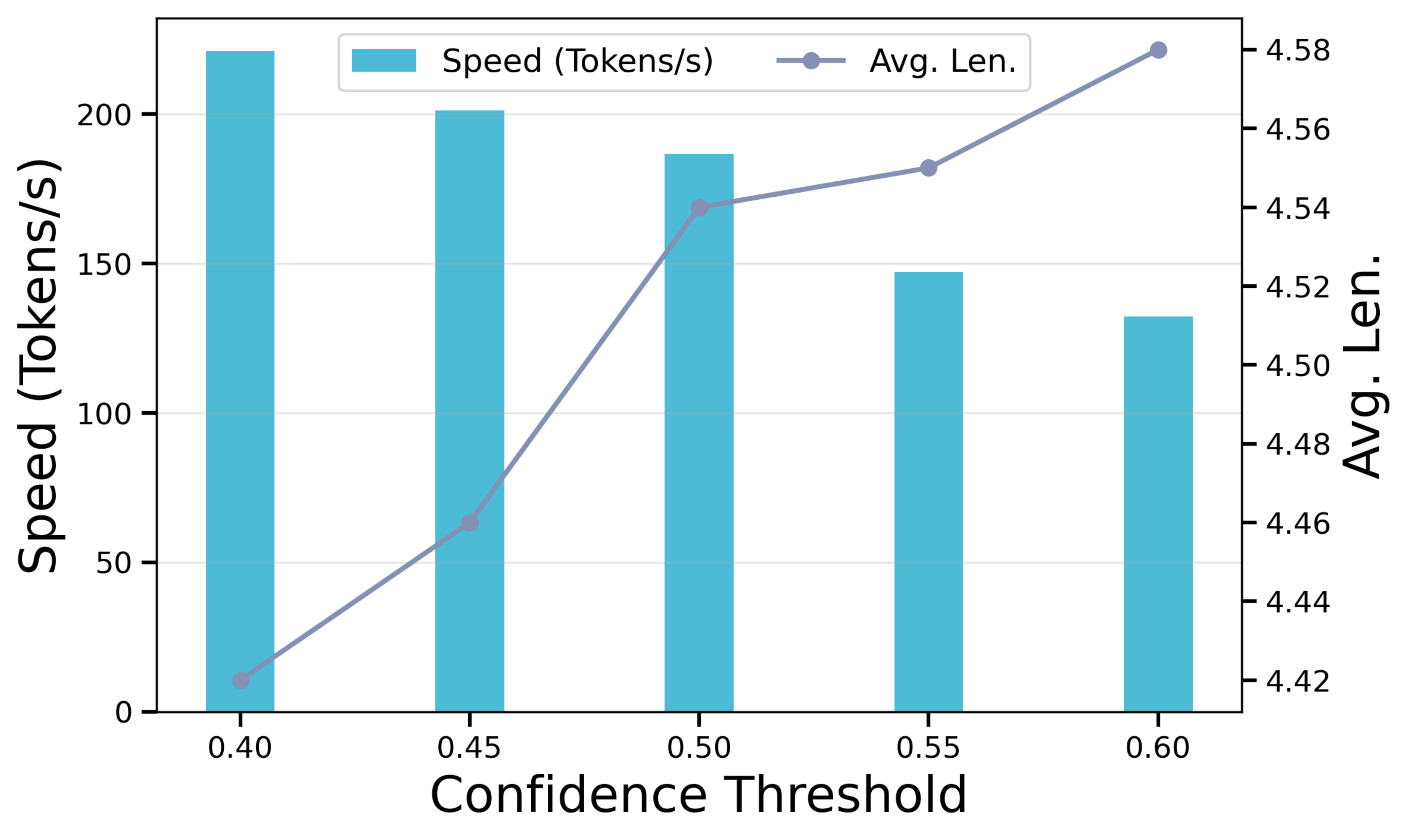} 
    \caption{
    % \textbf{example.} 
    % 加速效果和performance关于置信度的消融在UD-VLA上的消融，我们发现当confidence=0.5时，性能趋于稳定，并且此时加速比达到2.78\time,
    Ablation study on the confidence threshold~$\tau_{\text{conf}}$ of \method~based on UD-VLA.
    % shows the trade-off between acceleration and performance.
    % We choose a confidence threshold of 0.5 for the trade-off.
    % , at which performance stabilizes while achieving a speedup of $2.78\times$.
    }
    \label{fig:abl_conf} 
    % \vspace{-1.6cm}
\end{figure}

\noindent
\textbf{Ablation of $\tau_{\text{conf}}$.}
For the confidence threshold $\tau_{\text{conf}}$ in the semi-activated block, lowering the threshold yields an approximately linear drop in performance while improving inference speed. 
As shown in \Cref{fig:abl_conf}, we set the confidence threshold to 0.5 to balance these two factors, achieving a $2.8\times$ acceleration while incurring a marginal performance drop of 2\%.
\section{Related Works}
% \paragraph{Discrete VLA.}
\begin{wraptable}{r}{0.4\linewidth}
\vspace{-0.5cm}
\small
\centering
\caption{Choice of block size on LIBERO-Long based on Dream-VLA. Here, multiples denotes that the value is the multiples of the dimensionality of action, while random denotes random numbers for the block size. }
\label{tab:blocksize_libero}
\renewcommand{\arraystretch}{0.95}
\begin{tabular}{c c c}
\toprule
Block & Success Rate & Speedup \\
\midrule
Multiples   & 74.7\% & 4.01$\times$  \\
  Ramdom   & 73.3\% & 3.95$\times$  \\
\bottomrule
\end{tabular}
\vspace{-0.2cm}
\end{wraptable}
\paragraph{Discrete Diffusion VLA (dVLA).}
% The survey~\citep{yu2025discrete} defines dLLMs as models that adopt a multi-token, parallel decoding paradigm with full attention and perform generation via an iterative denoising process.
In this paper, we extend the notion of dLLM~\citep{yu2025discrete} to the embodied domain and define dVLA, a VLA model that enables parallel decoding of multiple action tokens (and optionally tokens from other modalities) through an iterative denoising-based inference procedure.
% and define dVLA, a Vision-Language-Action model that enables parallel decoding of multiple action tokens (and optionally tokens from other modalities) through an iterative denoising-based inference procedure.
PD-VLA~\citep{song2025accelerating} first adopts Jacobi decoding to enable AR VLA to predict in parallel action tokens without training.
% a BART-style~\citep{bart} denoising strategy. 
% In this approach, a subset of action tokens is randomly substituted with vocabulary tokens and then iteratively refined to recover the ground-truth sequence.
Then, DD-VLA~\citep{liang2025discrete} and LLADA-VLA~\citep{wen2025llada} follow the BERT-style~\citep{bert} masked prediction strategy, where selected action tokens are replaced with a special mask token, and the model directly learns to predict the original tokens at these masked positions. 
Dream-VLA~\citep{yedreamVLA} performs a large-scale robotic pretraining on a diffusion vision-language model to inject embodied capabilities.
% the cross-embodiment OXE dataset. 
UD-VLA~\citep{chen2025unified}, MM-ACT~\citep{liang2025mm} and dVLA~\citep{wen2025dvla} integrate visual CoT or textual CoT into discrete diffusion–based VLA models and jointly diffuse future frames, textual reasoning traces, and actions within a single unified framework.
However, they overlook the bottleneck in inference speed, thereby leaving a gap in real-world applications.
%然而，他们忽略了推理速度的瓶颈，从而与real world应用产生差距。
% To further improve efficiency, CEED-VLA~\citep{song2025ceed} employs consistency distillation to reduce the number of iterations in the discrete diffusion process.
% udvla,dvla,mmact将视觉cot或者文本cot融入到离散扩散vla中，共同扩散生成未来帧，文本cot以及动作，以离散扩散的方式统一了理解生成以及动作预测
% This pretraining scheme yields consistent performance gains for both downstream continuous diffusion and discrete diffusion fine-tuning settings.

% 但是这些文章都没有涉及缓存
% However, these discrete diffusion VLAs focus exclusively on action prediction while largely ignoring the interplay between visual and action tokens, thus failing to fully exploit the potential benefits of cross-modal representation learning.

\paragraph{Acceleration of VLA.}
% Various acceleration strategies, including quantization~\citep{lin2024awq} and token pruning~\citep{fastv}, have been effectively applied to LLMs, yet they often fail to meet the stringent real-time requirements of action generation. 
% Fast ECoT将异步推理理念扩展至具身思维链（ECoT）推理框架，通过策略性解耦潜推理与动作流，支持推理优化与输出生成并行，在不增加时间开销的前提下提升推理深度
% Spec-VLA首次将推测解码引入VLAs，并结合松弛接受机制提升草稿令牌接受率与平均长度，实现显著推理加速
% LightVLA引入自适应、性能导向的视觉令牌剪枝框架
% CronusVLA通过先进先出（FIFO）队列首创特征级令牌缓存，存储并复用紧凑运动特征，将计算密集的单帧感知与轻量化多帧推理解耦
% SmolVLA应用层剪枝，同时通过像素重排操作强制空间精简，run on consumer-grade hardware with only public datasets

% A growing body of work seeks to improve the efficiency of VLAs through both architectural and inference-time innovations. These efforts can be broadly categorized into several key approaches.
% One dominant strategy is pruning, which aims to eliminate redundant computations. For instance, EfficientVLA~\citep{yang2025efficientvla} introduces a systematic framework to remove unnecessary operations across the whole architecture. 
% More specialized techniques include ADP~\citep{pei2025action}, which enables action-aware dynamic token pruning, and LightVLA~\citep{jiang2025better}, which adopts an adaptive, performance-aware visual token pruning strategy. 
% Furthering this, MoLe-VLA~\citep{zhang2025mole} conditionally bypasses transformer layers based on task-relevant signals to reduce computation even more.
Recent efforts of efficient VLA focus on pruning for redundancy reduction. MoLe-VLA~\citep{zhang2025mole} dynamically activates layers via a Mixture-of-Layers design. EfficientVLA~\citep{yang2025efficientvla} proposes a training-free acceleration framework combining layer pruning, token selection, and diffusion caching.
ADP~\citep{pei2025action} and LightVLA~\citep{jiang2025better} introduce action-aware and differentiable token pruning strategies, respectively, to reduce visual redundancy while maintaining performance.
% DeeR-VLA~\citep{DeeR-VLA} ceed-vla 分别在模型深度，以及迭代次数上实现early exit 减少额外的计算
% For early exit, DeeR-VLA~\citep{DeeR-VLA} dynamically adapts the model depth during inference.
Beyond pruning, early-exit strategies have also been explored to reduce cost. DeeR-VLA~\citep{DeeR-VLA} adaptively adjusts the effective model depth, while CEED-VLA~\citep{song2025ceed} enables early termination over iterative steps during inference.
Another effective method is caching. VLA-Cache~\citep{vlacache}, for example, improves efficiency by caching static tokens and recomputing only task-dependent components. 
Similarly, CronusVLA~\citep{li2025cronusvla} proposes feature-level token caching via a FIFO queue, decoupling expensive single-frame perception from lightweight multi-frame reasoning.
Researchers are also exploring novel architectures and optimization techniques. RoboMamba~\citep{liu2024robomamba} integrates Mamba state-space modeling into the VLA framework to achieve efficient robotic reasoning. In parallel, quantization techniques, such as those proposed by BitVLA~\citep{wang2025bitvla} and QVLA~\citep{xu2026qvla}, enable the deployment of VLA models on hardware with limited resources by using low-bit representations.
Finally, the development of lightweight backbones from the ground up offers a direct path to efficiency. TinyVLA~\citep{wen2024tinyvla} pursues this by designing compact architectures from scratch. Flower~\citep{reuss2025flower} proposes an efficient 950M-parameter diffusion-based VLA that uses intermediate-modality fusion and action-specific conditioning. Meanwhile, SmolVLA~\citep{shukor2025smolvla} combines application-level pruning with spatial compaction through pixel rearrangement. 
However, these works do not specifically address the acceleration of dVLA. 
In contrast, our work systematically investigates acceleration strategies tailored to the unique characteristics of dVLA, thereby filling this gap.
% 但是这些方法并没有针对dVLA进行加速研究和适配,this work针对dVLA的特性进行加速研究，填补这一空白
% successfully enabling deployment on consumer-grade hardware using only public datasets.
\section{Conclusion}
% In this paper, we present \method, a novel AR--diffusion hybrid framework that accelerates dVLA models to real-time performance. 
% We introduce block-wise causal attention to enable stable KV caching, a progressively decaying noise schedule to unlock inter-block parallelism, and an asymmetric distillation strategy to efficiently transfer knowledge from pretrained bidirectional dVLAs.
% Extensive experiments across CALVIN, LIBERO, SIMPLER, and real-world benchmarks demonstrate consistent 2.8$\times$--4.1$\times$ acceleration over native dVLA models, while maintaining or even improving execution performance. 
% Importantly, \method requires only a small fraction of the original fine-tuning budget, making it an effective and economical post-training solution for upgrading existing dVLA models.

In this paper, we tackle key limitations in the inference speed of dVLAs.
Specifically, we reveal an implicit block-wise AR decoding tendency in the fully bidirectional dVLA. 
Thus, we propose \method, which leverages block-wise diffusion with a corresponding attention pattern to allow KV cache reuse, while allowing inter-block parallelism through diffusion forcing. 
We also curate an efficient training process and a pipelined inference for real-time inference.
Extensive experiments on simulated benchmarks and real-world tasks demonstrate up to 4.1$\times$ acceleration over existing dVLA models, while maintaining SOTA-level success rates.
These findings offer a practical solution for
deploying dVLAs as competitive alternatives to continuous flow-matching VLAs in real-world applications.
% Moreover, the results in diverse real-world tasks demonstrate the dynamic capability and working efficiency in the application.

% \section{Acknowledgments}

% \clearpage
% \newpage
\bibliographystyle{assets/plainnat}
\bibliography{paper}

\clearpage
\newpage
\onecolumn
\beginappendix
\renewcommand{\thefigure}{S\arabic{figure}}
\renewcommand{\thetable}{S\arabic{table}}
\setcounter{figure}{0}
\setcounter{table}{0}

% \section{Cross-entropy Training from dVLA}  
% This method directly optimizes the model under the block-diffusion objective without requiring a teacher diffusion model.
% \begin{equation}
% \label{eq:bd_loss}
% \mathcal{L}_{\text{BD}}
% = \mathbb{E}\sum_{i=1}^{N} 
% \left[
%     -\log p_{\theta}(Y_{B_i}^0 |Y_{B_<i}^{t_<i},c)
% \right].
% \end{equation}

% \Cref{fig:training_step} shows that $\mathcal{L}_{\text{AD}}$ achieves convergence with only one-tenth of the steps required for full fine-tuning, while $\mathcal{L}_{\text{BD}}$ requires nearly half of the fine-tuning budget. Unless otherwise stated, we adopt $\mathcal{L}_{\text{AD}}$ as the default training objective.
\noindent
This supplementary material provides additional analyses and implementation details for \method. We first present ablation studies on the decoding hyperparameters, including the block expansion and activation thresholds in \Cref{sec:tau_ablation} and the radical decoding strategy in fully activated blocks in \Cref{sec:radical_ablation}. We then describe the implementation details in \Cref{sec:impl_details} and provide the detailed inference procedure in \Cref{sec:inference_details}. Next, we summarize the evaluation benchmarks in \Cref{sec:benchmark}, followed by the comparison with state-of-the-art methods on LIBERO in \Cref{sec:libero_sota}.

\begin{table}[t]
\centering
\small
\setlength{\tabcolsep}{4pt}
\renewcommand{\arraystretch}{0.8}
\caption{Ablation study on block expansion threshold $\tau_{\text{add}}$ and block activation threshold $\tau_{\text{act}}$ on UD-VLA. }
\label{tab:threshold_ablation}
\begin{tabular}{lccc}
\toprule
$\tau_{\text{add}}$ & $\tau_{\text{act}}$ & avg. len. $\uparrow$ & Speed $\uparrow$  \\
\midrule
0.4 & 0.4 & 4.42 & \textbf{253.1}  \\
0.4 & 0.6 & 4.46 & 248.3  \\
0.5 & 0.5 & 4.44 & 204.8  \\
0.5 & 0.7 & 4.54 & 186.7  \\
0.6 & 0.6 & 4.52 & 182.3  \\
0.6 & 0.8 & \textbf{4.57} & 160.7  \\
\bottomrule
\end{tabular}
\end{table}
\section{Ablation of $\tau_{\text{add}}$ and $\tau_{\text{act}}$.}
\label{sec:tau_ablation}
%我们基于ud-vla 对 $\tau_{\text{add}}$ and $\tau_{\text{act}} \cref{} 分析，我们保持block——size=64  $\tau_{\text{conf}} 为0。5 ，$\tau_{\text{add}}$ and $\tau_{\text{act}} 相等时，块一被添加就变成full-activate，双状态解码退化成单状态，当table$shows that $\tau_{\text{add}}$ and $\tau_{\text{act}}大致上与性能成正比的趋势，与速度成相反的趋势，同时双状态解码会较少性能损失，同时速度与单状态几乎一致，对于action对正确率要求高的场景，我们建议使用更保守的双状态解码（\tau_{\text{add}}$ > $\tau_{\text{act}}$）
% \textbf{Ablation of $\tau_{\text{add}}$ and $\tau_{\text{act}}$.}
We conduct an ablation study on $\tau_{\text{add}}$ and $\tau_{\text{act}}$ based on UD-VLA (see \Cref{tab:threshold_ablation}). 
% Throughout the experiments, we fix the block size to 64 and set the confidence threshold $\tau_{\text{conf}}$ to 0.5.
When $\tau_{\text{add}} = \tau_{\text{act}}$, a newly added block immediately becomes fully activated, causing the dual-state decoding scheme to degenerate into a single-state regime.
The results show that our~\method~is not sensitive to these hyperparameters. 
$\tau_{\text{add}}$ and $\tau_{\text{act}}$ exhibit a positive correlation with task performance and a negative correlation with decoding speed.
Importantly, the proposed dual-state decoding mechanism incurs significantly performance degradation while maintaining decoding speed comparable to the single-state variant.
For scenarios that demand high action accuracy, we adopt a more conservative dual-state configuration (i.e., $\tau_{\text{add}} < \tau_{\text{act}}$) to better preserve performance.

\begin{table*}[t]
\setlength{\tabcolsep}{4pt}
\renewcommand{\arraystretch}{0.9}
\centering
\small
\caption{Comparison with various base models in terms of inference speed, average length, and execution frequency.}
\label{tab:baselines}

\begin{tabular}{@{} c  c c  @{}}
\toprule
Radical Decoding
& Speed (Tokens/s) 
& Avg. Len. \\
\midrule

log2
& 186.67 
& 4.54  \\

log3 
& 164.42
& 4.57  \\

log4
& 144.71 
& 4.58     \\

\bottomrule
\end{tabular}

\vspace{-0.2cm}
\end{table*}

\section{Ablation Study of Radical Decoding in Fully Activated Blocks}

\label{sec:radical_ablation}

We further conduct an ablation study on the radical decoding strategy used in fully activated blocks. Specifically, we vary the radical decoding factor among \texttt{log2}, \texttt{log3}, and \texttt{log4}. Here, \texttt{log2} corresponds to the most aggressive setting, where a fully activated block decodes at least half of its remaining tokens in one iteration.

As shown in Table~\ref{tab:baselines}, our~\method~is fairly robust to different radical decoding factors. Although more conservative settings such as \texttt{log3} and \texttt{log4} slightly reduce the decoding speed, the average success length remains largely stable across all configurations. In particular, \texttt{log2} achieves the highest decoding speed of 186.67 tokens/s, while maintaining a comparable average success length of 4.54. 
These results suggest that our~\method~is not highly sensitive to the choice of the radical decoding factor, and that the aggressive \texttt{log2} setting provides the best efficiency-performance trade-off.

\section{Implementation Details.}
\label{sec:impl_details}
We employ LoRA-based asymmetric distillation throughout. The LoRA rank is set to 32. During distillation, LoRA branches are disabled when computing teacher output logits. In contrast, LoRA branches are activated when computing student model output logits. This design maximally preserves the pretrained dVLA backbone’s visual–language understanding and action-reasoning priors, while allowing the LoRA modules to focus solely on learning the transfer of attention pattern.

For both Dream-VLA and DD-VLA, we set the action chunk size to 8, and to 5 for the SIMPLER tasks. For UD-VLA, the action chunk size is set to 10. All other training hyperparameters follow the official settings.

\section{Inference Details.}
\label{sec:inference_details}
% 详细的推理伪代码 as shown in \Cref{alg:block_conf_log}
The detailed inference pseudocode is provided in \Cref{alg:block_conf_log}.
\begin{algorithm}[t]
\footnotesize
\caption{Confidence-Guided Block Decoding with Logarithmic Scheduling}
\label{alg:block_conf_log}
\begin{algorithmic}[1]
\footnotesize
\REQUIRE Fast-dVLA model $p_{\theta}$; 
block expansion threshold $\tau_{\text{add}}$; 
block activation threshold $\tau_{\text{act}}$; 
confidence threshold $\tau_{\text{conf}}$; 
logarithmic scheduling factor $n$.

\STATE Initialize $Y=\{Y_{B_1}\}$ as a single block filled with \texttt{[MASK]} tokens.

\WHILE{generation is not complete}

    \IF{the decoded ratio in $Y_{B_{i-1}}$ exceeds $\tau_{\text{add}}$ \AND \texttt{<|EOA|>} not in $Y$}
        \STATE Append a new block $Y_{B_i}$ with all tokens masked and mark it as \emph{semi-activated}.
    \ENDIF

    \STATE Perform a forward pass on $Y$ using Fast-dVLA $p_{\theta}$ with cached KV.

    \FOR{each active block $Y_{B_i}$ in $Y$}

        \STATE Let $\mathcal{R}_i$ denote the set of remaining masked token positions in $B_i$.
        \STATE Compute confidence scores $\mathbf{c}_i$ for all positions in $\mathcal{R}_i$.

        \IF{$Y_{B_i}$ is \emph{fully activated}}
            \STATE Set $k \leftarrow \lfloor |\mathcal{R}_i| / n \rfloor$.
            \STATE Compute the block-specific decoding threshold: \\
            $
            \tau_i \leftarrow \min\!\big(\tau_{\text{conf}},\ \min(\mathrm{TopK}(\mathbf{c}_i, k))\big).
            $
        \ELSE
            \STATE Set $\tau_i \leftarrow \tau_{\text{conf}}$.
        \ENDIF

        \STATE Construct the decoding candidate set
        $
        \mathcal{S}_i \leftarrow \{\, p \in \mathcal{R}_i \mid c_i(p) \ge \tau_i \,\}.
        $

        \STATE Decode tokens at positions in $\mathcal{S}_i$ and remain others as mask token in $B_i$.

        \IF{the decoded ratio in $Y_{B_{i-1}}$ exceeds $\tau_{\text{act}}$}
            \STATE Mark $Y_{B_i}$ as \emph{fully activated}.
        \ENDIF

    \ENDFOR

    \STATE Update the KV cache for completed blocks.

\ENDWHILE

\end{algorithmic}
\end{algorithm}

\section{Benchmarks}
\label{sec:benchmark}
\textbf{CALVIN.}
The CALVIN benchmark~\citep{mees2022calvin} is a simulated suite for evaluating long-horizon, language-conditioned robotic manipulation. It spans four environments (A, B, C, and D) with 34 tasks and 1,000 language instructions. We evaluate 500 rollouts per model, where each rollout involves a sequence of 5 consecutive sub-tasks. We report the average length (avg. len.) of successful sub-task completions of all rollouts with a maximum value of 5.

\noindent
\textbf{LIBERO.}
% The LIBERO benchmark~\citep{liu2023libero} is a simulation suite for robotic manipulation, comprising four task suites: Spatial, Object, Goal, and Long.
% Each suite has 10 tasks and 50 human-teleoperated demonstrations. LIBERO-Spatial evaluates spatial reasoning with varied layouts, while LIBERO-Object tests object-level generalization across different objects. LIBERO-Goal examines goal-conditioned behavior, and LIBERO-Long presents long-horizon, compositional tasks that challenge temporal and compositional reasoning. 
% We report the success rates for all four task suites as well as the overall average. 
% Each suite is evaluated over 500 rollouts per task.
LIBERO~\citep{liu2023libero} is a simulated manipulation benchmark with 4 suites (Spatial, Object, Goal, Long). Spatial probes layout reasoning, Object tests object generalization, Goal evaluates goal-conditioned control, and Long targets long-horizon compositional skills. We report success rates per suite and overall average, each suite containing 10 tasks and 50 rollouts per task.
% 我们report四个类别任务的成功率以及总成功率，每个task suites测试500次
% Overall, LIBERO provides a comprehensive testbed for evaluating diverse generalization abilities in robotic manipulation.

\noindent
\textbf{SimplerEnv.}
SimplerEnv~\citep{li24simpler} is a real-to-sim suite for assessing transfer and generalization of robot policies trained on real-world video data. We evaluate on WidowX robots under varied lighting, textures, colors, and viewpoints. Tasks include \textit{Put Spoon on Towel}, \textit{Put Carrot on Plate}, \textit{Stack Green on Yellow Block}, and \textit{Put Eggplant in Yellow Basket}. We report per-task success rates and the overall average.

% \section{Implementation Details}
\begin{table*}[t]
\setlength{\tabcolsep}{10pt}
\footnotesize
\centering
\caption{\textbf{Evaluation and comparison on the LIBERO benchmark.}}
\begin{tabular}{lccccc}
\toprule
\textbf{Method} & \textbf{Spatial} & \textbf{Object} & \textbf{Goal} & \textbf{Long} & \textbf{Average} \\
\midrule

Octo~\citep{octo_2023}               
& 78.9\% & 85.7\% & 84.6\% & 51.1\% & 75.1\% \\

SpatialVLA~\citep{qu2025spatialvla}  
& 88.2\% & 89.9\% & 78.6\% & 55.5\% & 78.1\% \\

CoT-VLA~\citep{zhao2025cot}          
& 87.5\% & 91.6\% & 87.6\% & 69.0\% & 81.1\% \\

WorldVLA~\citep{worldvla}          
& 87.6\% & 96.2\% & 83.4\% & 60.0\% & 81.8\% \\

ThinkAct~\citep{huang2025thinkact}          
& 88.3\% & 91.4\% & 87.1\% & 70.9\% & 84.4\% \\

$\pi_0$-FAST~\citep{pertsch2025fast} 
& 96.4\% & 96.8\% & 88.6\% & 60.2\% & 85.5\% \\

MolmoAct~\citep{lee2025molmoact}          
& 87.0\% & 95.4\% & 87.6\% & 77.2\% & 86.6\% \\

FlowVLA~\citep{zhong2025flowvla}          
& 93.2\% & 95.0\% & 91.6\% & 72.6\% & 88.1\% \\

DreamVLA~\citep{dreamvla25}          
& \underline{97.5\%} & 94.0\% & 89.5\% & 89.5\% & 92.6\% \\

$\pi_0$~\citep{Pi0}          
& 96.8\% & \textbf{98.8\%} & 95.8\% & 85.2\% & 94.2\% \\

$\pi_{0.5}$~\citep{intelligence2025pi_}          
& \textbf{98.8\%} & 98.2\% & \textbf{98.0\%} & \underline{92.4\%} & \textbf{96.8\%} \\

DDVLA~\citep{liang2025discrete}
& 97.2\% & 98.6\% & 97.4\% & 92.0\% & 96.3\% \\

\rowcolor[gray]{0.9} \hspace{3mm} + ours 
& 97.0\% & \textbf{98.8\%} & \underline{97.6\%} & \textbf{92.8\%} & \underline{96.6\% }\\
\bottomrule
\end{tabular}
\label{tab:libero}
\vspace{-3mm}
\end{table*}
\section{Comparison with SOTA on LIBERO}
\label{sec:libero_sota}

Table~\ref{tab:libero} presents the comparison with state-of-the-art methods on the LIBERO benchmark. Overall, our method achieves competitive performance against recent strong VLA baselines, demonstrating that accelerating discrete diffusion VLAs does not compromise policy quality. Built upon DDVLA~\citep{liang2025discrete}, our method improves the average success rate from 96.3\% to 96.6\%, while further boosting performance on the most challenging \textit{Long} suite from 92.0\% to 92.8\%. The gains on long-horizon tasks suggest that our acceleration strategy preserves, and even slightly enhances, the sequential decision-making capability required for extended manipulation.

Compared with previous autoregressive and continuous-flow paradigms, our method remains highly competitive across all four suites. In particular, it matches or surpasses strong baselines such as $\pi_0$-FAST, MolmoAct, and FlowVLA by a clear margin, and performs comparably to the most recent frontier models, including $\pi_0$, $\pi_{0.5}$, and DDVLA. Notably, while $\pi_{0.5}$ achieves the best average success rate, our \method~delivers stronger performance than DDVLA on \textit{Object}, \textit{Goal}, and especially \textit{Long}, highlighting the effectiveness of our design in improving both robustness and long-horizon execution.

These results indicate that our method inherits the strong representational capacity of discrete diffusion VLAs, while making them more practical without sacrificing task success.

\newpage

\end{document}